\documentclass{article}

\usepackage[main,final,nonatbib]{neurips_2026}

\usepackage[utf8]{inputenc} % allow utf-8 input
\usepackage[T1]{fontenc}    % use 8-bit T1 fonts
\usepackage{hyperref}       % hyperlinks
\usepackage{url}            % simple URL typesetting
\usepackage{booktabs}       % professional-quality tables
\usepackage{amsfonts}       % blackboard math symbols
\usepackage{nicefrac}       % compact symbols for 1/2, etc.
\usepackage{microtype}      % microtypography
\usepackage{xcolor}         % colors
\usepackage{booktabs}    % 
\usepackage{tablefootnote} % 
\usepackage{multirow}  %
 \usepackage{graphicx}
\usepackage[utf8]{inputenc}
\usepackage{svg}
\usepackage{tabularx}
\usepackage{booktabs}
\usepackage{multirow}
\usepackage{makecell}
\usepackage{graphicx}

\usepackage[table]{xcolor} 
\usepackage{booktabs}
\usepackage{multirow}
\usepackage{graphicx}
\usepackage{makecell}
\usepackage{graphicx}
\usepackage{amssymb} 
\usepackage{bbm}
\usepackage{booktabs}
\usepackage{amsmath}
\usepackage[utf8]{inputenc}
\usepackage[ruled,linesnumbered]{algorithm2e}
\usepackage{xcolor}
\usepackage[utf8]{inputenc}
\usepackage{geometry}
\usepackage{graphicx}
\usepackage{makecell}
\usepackage{amsmath}
\definecolor{commentColor}{RGB}{34,139,34}
\usepackage{array} % 用于自定义列格式\usepackage{booktabs} % 用于 \toprule, \midrule, \bottomrule\usepackage{array}
\usepackage{wrapfig}
\title{DrawingsDreamer: A Unified Multi-View Engineering Drawings Generation Model}

\author{
Shurui Liu$~~^{1}$ ~~
Weide Chen$^{2}$ ~~
Changwang Yi$^{1}$ ~~~
Ancong Wu$^{1}$\thanks{~Corresponding author.} ~~~
\\
$^1$ School of Computer Science and Engineering, Sun Yat-sen University, China  \\
$^2$ School of Intelligent Systems Engineering, Shenzhen Campus of Sun Yat-sen University, China  \\
\texttt{\{liushr29, chenwd56, yichw\}@mail2.sysu.edu.cn}\\
\texttt{wuanc@mail.sysu.edu.cn}
\vspace{-0.4cm} 
}

\begin{document}

\maketitle

\begin{figure}[htbp]
    \centering
    \includegraphics[width=\linewidth]{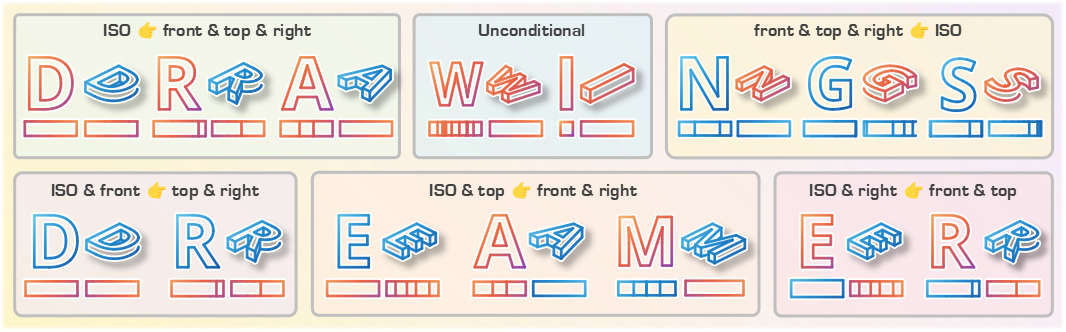}
\caption{\textbf{DrawingsDreamer} handles a full spectrum of multi-view modeling tasks. \textbf{The 
% orange gradients 逐字渐变（橙色 -> 粉紫色）
\textcolor{orange!100!magenta}{o}\textcolor{orange!93!magenta}{r}\textcolor{orange!85!magenta}{a}\textcolor{orange!78!magenta}{n}\textcolor{orange!71!magenta}{g}\textcolor{orange!63!magenta}{e} %
\textcolor{orange!45!magenta}{g}\textcolor{orange!38!magenta}{r}\textcolor{orange!30!magenta}{a}\textcolor{orange!23!magenta}{d}\textcolor{orange!15!magenta}{i}\textcolor{orange!8!magenta}{e}\textcolor{orange!0!magenta}{n}\textcolor{orange!0!magenta}{t}\textcolor{orange!0!magenta}{s}
denote the outputs, while the 
% blue gradients 逐字渐变（浅蓝 -> 深蓝）
\textcolor{cyan!100!blue}{b}\textcolor{cyan!92!blue}{l}\textcolor{cyan!85!blue}{u}\textcolor{cyan!77!blue}{e} %
\textcolor{cyan!62!blue}{g}\textcolor{cyan!54!blue}{r}\textcolor{cyan!46!blue}{a}\textcolor{cyan!38!blue}{d}\textcolor{cyan!31!blue}{i}\textcolor{cyan!23!blue}{e}\textcolor{cyan!15!blue}{n}\textcolor{cyan!8!blue}{t}\textcolor{cyan!0!blue}{s}
represent the inputs.} In each example, the multi-view layout arranges the Top, Front, and Right orthographic views at the top-left, bottom-left, and bottom-right, respectively, with the Isometric view situated at the top-right.}
    \label{fig:teaser}
\end{figure}

\begin{abstract}
Scalable Vector Graphics (SVG) are essential for modern industrial Computer-Aided Design (CAD). However, existing autoregressive SVG generation models are predominantly tailored for artistic creation and struggle to maintain the rigorous geometric fidelity and cross-view spatial alignment required for engineering drawings. To bridge this gap, we introduce \textbf{DrawingsDreamer}, a unified Large Language Model (LLM)-driven framework for multi-view vector-based engineering drawings generation. By formulating the generation of multi-view engineering drawings purely as a sequence modeling task, we eliminate the need of raster image encoders. We propose a Streamlined Representation utilizing hierarchical postfix tokenization, which guides the model to establish local geometric coordinates before assigning semantic boundaries. Optimized via a progressive task-aware curriculum schedule, \textbf{DrawingsDreamer} effectively transitions from localized structural repair to macroscopic generation in a unified model. Extensive experiments demonstrate that our unified model achieves strong performance in both geometric fidelity and syntactic accuracy across diverse conditional and unconditional generation tasks.
\end{abstract}

\section{Introduction}

Scalable Vector Graphics (SVG) have long served as the \textit{de facto} standard in modern digital design, spanning applications from user interfaces (UI) to complex industrial Computer-Aided Design (CAD) systems. Their widespread adoption is driven by resolution independence, compact file sizes, and the capacity for precise, parametric control over geometric primitives such as polygons and Bézier curves. While recent advancements in generative AI have revolutionized visual synthesis, the creation of highly structured, multi-view engineering drawings remains a formidable challenge. As the foundation of modern manufacturing, engineering drawings demand strict structural validity, exact geometric parameterization, and rigorous cross-view spatial alignment. These qualities differ fundamentally from the requirements of general artistic creation.

Existing approaches to SVG generation largely fall short of these industrial demands. Optimization-based methods~\cite{ma2022towards,li2020differentiable} often struggle with prohibitive computational overhead and tend to produce unstructured, dense anchor points that destroy the editability inherent to the SVG format. Conversely, autoregressive methods and LLM-driven pipelines~\cite{touvron2023llama,achiam2023gpt,yang2025qwen3} have demonstrated remarkable scalability by treating SVG generation as a sequence modeling task. However, bottlenecked by limited context lengths and training corpora overwhelmingly skewed towards artistic domains (e.g., icons, fonts, and illustrations)~\cite{xing2025empowering,rodriguez2025starvector}, these models fail to capture the long-range coordinate dependencies and spatial constraints required for engineering CAD drafts. Concurrently, the broader CAD generation community has predominantly focused on synthesizing 3D Boundary Representations (BRep) or meshes from point clouds and images~\cite{wu2021deepcad,xu2024brepgen}, leaving the direct generation of the 2D drawings themselves entirely unexplored.

Aligning with the broader consensus that scalable, general-purpose sequence modeling ultimately outpaces hand-crafted optimization heuristics, we frame the synthesis of 3D engineering drawings purely as a 2D sequence generation paradigm. In this paper, we propose \textbf{DrawingsDreamer}, the first unified multi-view engineering drawings generation model. By projecting 3D CAD objects into a canonical set of principal views, we deliberately avoid harnessing any raster image encoders. Instead, we rely entirely on the robust spatial reasoning capabilities of LLMs operating over discrete coordinate tokens. To effectively map continuous geometric spaces to discrete vocabularies, we introduce a \textit{Streamlined Representation} featuring a novel hierarchical postfix tokenization strategy. This formulation compels the model to robustly establish continuous local coordinates prior to encapsulating them with semantic boundaries, thereby stabilizing the generation of complex spatial structures.

To facilitate the training of our model, we systematically construct a multi-view aligned engineering drawings corpus by rendering standard projections from existing 3D CAD collections. Leveraging this structured data, we optimize  \textbf{DrawingsDreamer} through a progressive, task-aware curriculum schedule, empowering a single unified model to seamlessly transition between microscopic topological repair, partial view completion, and unconstrained macroscopic drawings imagination.

In summary, our core contributions are as follows:

\begin{itemize}
\item We propose a \textit{Streamlined Representation} utilizing a hierarchical postfix tokenization strategy. This formulation effectively maps continuous geometric spaces into a discrete, LLM-friendly vocabulary, fundamentally stabilizing the autoregressive generation of complex spatial coordinates.
\item We introduce \textbf{DrawingsDreamer}, the first unified framework capable of generating highly structured, multi-view parametric engineering drawings. By optimizing the model via a progressive, task-aware curriculum schedule, we empower a single framework to seamlessly transition between microscopic topological repair and macroscopic spatial generation.
\item Extensive evaluations demonstrate that \textbf{DrawingsDreamer} achieves state-of-the-art performance in both geometric fidelity and syntactic accuracy across diverse conditional and unconditional multi-view generation tasks.
\end{itemize}

\section{Related Work}
\subsection{SVG Generation}
The automated generation of vector graphics has witnessed a significant paradigm shift from instance-level optimization to large-scale sequence modeling. Early optimization methods~\cite{ma2022towards,li2020differentiable} iteratively refine SVG parameters by minimizing photometric loss via differentiable rasterizers. However, these approaches rely heavily on iterative optimization for individual instances. Consequently, they suffer from prohibitive computational overhead and often yield unstructured outputs lacking geometric fidelity. In contrast, driven by the shift towards large-scale data and general-purpose computation, learning-based SVG generation has witnessed significant progress. Advancements range from the curation of extensive vector datasets~\cite{wang2021deepvecfont,clouatre2019figr,kocetkov2022stack,yang2025omnisvg} to the development of generative pipelines built upon RNNs~\cite{song2023clipvg,ha2017neural,reddy2021im2vec,hu2024supersvg} and VAEs~\cite{carlier2020deepsvg,lopes2019learned,tang2024strokenuwa,su2023marvel,tian2022modern}. More recently, the paradigm has shifted towards treating SVG generation as a sequence modeling task by leveraging Large Language Models (LLMs)~\cite{touvron2023llama,achiam2023gpt,yang2025qwen3,das2025security,vaswani2017attention}, which demonstrate exceptional generative flexibility across diverse domains~\cite{li2023fine,zhang2025collm,shojaee2024llm,zhang2024llama,jiang2023motiongpt,gu2025effectiveness,dong2025codescore}.

Despite these advances, existing LLM-driven SVG generation models~\cite{xing2025empowering,rodriguez2025starvector,wu2025chat2svg,yang2025omnisvg} remain bottlenecked by limited context lengths and training corpora restricted to artistic or visual creation. Consequently, generating highly structured engineering drawings with aligned multiview representations remains an unexplored frontier, limiting the adoption in professional CAD workflows. To bridge this gap, we propose the first unified framework for generating consistent multiview parametric engineering drawings, effectively scaling these capabilities to complex industrial applications.

\subsection{Computer-Aided Design}
The domain of Computer-Aided Design (CAD) has rapidly evolved from foundational datasets~\cite{wu2021deepcad,koch2019abc,willis2021fusion} to sophisticated generative pipelines~\cite{li2025brepgpt,xu2025autobrep,li2025stitch,lee2025brepdiff,li2025revisiting,wu2024cadvlm,alrashedy2024generating,wu2021deepcad,xu2024brepgen,li2025dtgbrepgen,liu2026hidigen,khan2024cad,rukhovich2025cad,chen2025cadcrafter,li2025caddreamer,wang2025cadgpt,alrashedy2024generating,yavartanoo2024text2cad,khan2026dreamcad}. Current methods predominantly focus on reverse engineering, generating Boundary Representation (BRep) models from modalities like point clouds~\cite{wu2021deepcad,xu2024brepgen,li2025dtgbrepgen,liu2026hidigen,khan2024cad,rukhovich2025cad}, images~\cite{chen2025cadcrafter,li2025caddreamer}, or text~\cite{wang2025cadgpt,alrashedy2024generating,yavartanoo2024text2cad,khan2026dreamcad}. 

Furthermore, while some techniques reconstruct CAD models from static engineering drawings~\cite{gong2006reconstruction111,kuo1998reconstruction222,liu2001reconstruction333,wang1993survey444,furferi20102d555,camba2022sketch,harish2021photo2cad,puhachov2023reconstruction,wang20252d,zhang2023automatic,qin2025drawing2cad}, generating the drawings themselves remains entirely unexplored. As the fundamental blueprints of modern manufacturing, automating the creation of these drawings represents a critical missing link. To address this void and facilitate the training of generative models, we establish a scalable pipeline to derive structured, multi-view aligned 2D drawings directly from standard 3D CAD databases.
\section{Method} \label{sec: method}
In this section, we present the architecture and training pipeline of \textbf{DrawingsDreamer}.  We first introduce our \textit{Streamlined Representation} (Section~\ref{sec:representation}), which employs a hierarchical postfix tokenization scheme to effectively map continuous parametric geometries into a discrete, LLM-friendly vocabulary. Building upon this formulation, we then detail our unified instruction-tuning paradigm and progressive task-aware curriculum schedule (Section~\ref{sec:finetuning}). This dynamic optimization strategy empowers a single foundation model to seamlessly master both microscopic topological validity and rigorous cross-view spatial alignment.
\subsection{Streamlined Representation for Multi-View Engineering Drawings}
\label{sec:representation}

To effectively harness the scaling capabilities of LLMs for engineering drawings generation, we formulate the synthesis of engineering drawings as a pure sequence modeling paradigm. Specifically, a 3D CAD object $\mathcal{M}$ is projected into a canonical set of four principal views $\mathcal{V} = \{v_{\text{front}}, v_{\text{right}}, v_{\text{top}}, v_{\text{iso}}\}$. This projection provides a complete, unambiguous geometric description while deliberately avoiding the  computational overhead associated with rendering and processing rasterized images via visual encoders. We encode this geometry into a \textit{Streamlined Representation}, a compact training format optimized purely for LLM comprehension.

\begin{figure}
    \centering
    \includegraphics[width=\linewidth]{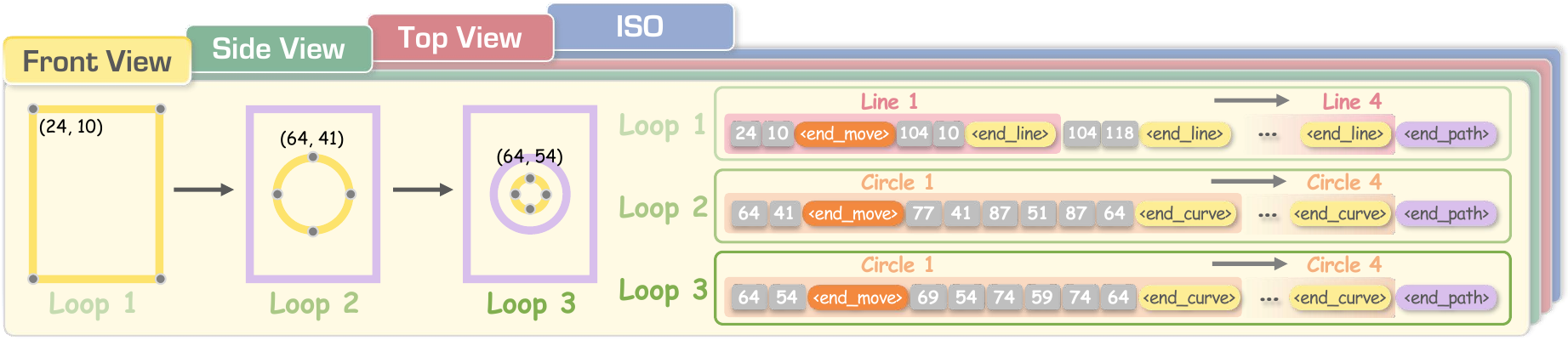}
\caption{Visualization of the Streamlined Representation, detailing its hierarchical data organization. Continuous coordinates are grouped into primitive boundaries (e.g., lines and curves), formed into closed loops, and ultimately structured into unified view-level sequences.}
\label{fig:representation}
\end{figure}

\paragraph{Autoregressive Postfix Tokenization.}
While prior works typically adopt a prefix-based tokenization scheme~\cite{yang2025omnisvg}, we introduce a hierarchical \textit{postfix tokenization} strategy. The model is compelled to first regress the  geometric coordinates, subsequently encapsulating them with a semantic boundary token (e.g., \texttt{<end\_curve>}).  To maintain a streamlined vocabulary and reduce the model token search space, the standard SVG close path command (\texttt{Z} or \texttt{z}) is not assigned a unique token. Instead, our parser explicitly converts it into a line sequence connecting the current position back to the subpath starting coordinate, terminated by \texttt{<end\_line>}. The geometric definitions of our structural vocabulary are summarized in Table~\ref{tab:vocabulary}.

\begin{table}[h]
\centering
\caption{Vocabulary of structural boundary tokens and their geometric semantics.}
\label{tab:vocabulary}
  \resizebox{0.9\textwidth}{!}{

\begin{tabular}{llp{8cm}}
\toprule
\textbf{Hierarchy Level} & \textbf{Token Format} & \textbf{Geometric Semantics} \\
\midrule
\multirow{3}{*}{\shortstack[l]{Primitive\\Boundaries}} 
& $x, y$ \texttt{<end\_move>} & \textbf{Move (\texttt{M}):} Initializes a new subpath at coordinate $(x, y)$. \\
& $x, y$ \texttt{<end\_line>} & \textbf{Line (\texttt{L}):} Draws a straight segment from the current position to $(x, y)$. \\
& $x_1, y_1, x_2, y_2, x, y$ \texttt{<end\_curve>} & \textbf{Cubic Bézier (\texttt{C}):} Draws a curve ending at $(x, y)$ using control points $(x_1, y_1)$ and $(x_2, y_2)$. \\
\midrule
\multirow{3}{*}{\shortstack[l]{Macro\\Boundaries}} 
& \texttt{<end\_path>} & \textbf{Path Boundary:} Terminates a continuous geometric path $\mathcal{P}$. \\
& \texttt{<end\_view>} & \textbf{View Boundary:} Terminates a spatial projection view sequence $S_v$. \\
& \texttt{<mask\_path>} & \textbf{Mask Placeholder:} Represents a masked path $\mathcal{P}_{\text{mask}} \in S_{\text{context}}$ during training. \\
\bottomrule
\end{tabular}
}
\end{table}

\paragraph{Coordinate Normalization.} 
To bridge continuous geometric spaces with discrete token vocabularies, coordinates are globally scaled to preserve cross-view spatial alignment (with the isometric view scaled independently) and quantized into $N$ discrete integer bins. The mathematical formulation is deferred to Appendix.

\paragraph{Unified Multitask Sequence Construction.} As visualized in Figure~\ref{fig:representation}, rather than statically concatenating all views, we serialize the quantized paths into a flexible layered sequence designed to support a unified multitask training framework. The sequence is built hierarchically: coordinates are grouped by primitive boundary tokens (e.g., \texttt{<end\_curve>}), primitives are grouped into paths terminated by \texttt{<end\_path>}, and paths are aggregated into a complete view sequence $S_v$, strictly terminated by its corresponding view boundary token (e.g., \texttt{<end\_front>}).

To empower the LLM with diverse generative capabilities, we dynamically construct the final drawings sequence as an instruction following prompt:
\begin{equation}
    S_{\mathcal{M}} = [\mathcal{I}, S_{\text{context}}, S_{\text{target}}] .
\end{equation}
Here, $\mathcal{I}$ acts as a task specific instruction. During training, we employ a curriculum learning strategy, dynamically sampling from a rich set of macroscopic generation and microscopic restoration tasks. To formalize these tasks, we define the orthographic view set as $\mathcal{V}_{\text{ortho}} = \{v_{\text{front}}, v_{\text{right}}, v_{\text{top}}\}$, and the complete view set as $\mathcal{V} = \mathcal{V}_{\text{ortho}} \cup \{v_{\text{iso}}\}$. The configurations of $S_{\text{context}}$ and $S_{\text{target}}$ are systematically varied across epochs as detailed in Table~\ref{tab:multitask_configs}.

\begin{table}[h]
\centering
\caption{Summary of unified multitask configurations. $\mathcal{V}_{\text{ortho}}$ denotes the set of orthographic views, and $v_i$ represents a single orthographic view.}
\label{tab:multitask_configs}
  \resizebox{0.8\textwidth}{!}{
\begin{tabular}{lll}
\toprule
\textbf{Task Configuration} & \textbf{Input Context ($S_{\text{context}}$)} & \textbf{Target Response ($S_{\text{target}}$)} \\
\midrule
Unconditional Generation & $\emptyset$ & $\mathcal{V}$ \\
Isometric to Orthographic & $\{v_{\text{iso}}\}$ & $\mathcal{V}_{\text{ortho}}$ \\
Orthographic to Isometric & $\mathcal{V}_{\text{ortho}}$ & $\{v_{\text{iso}}\}$ \\
Partial View Completion & $\{v_{\text{iso}}, v_i\} \mid v_i \in \mathcal{V}_{\text{ortho}}$ & $\mathcal{V}_{\text{ortho}} \setminus \{v_i\}$ \\
Path Level Semantic Masking & $S_{\text{available}} \setminus \{\mathcal{P}_{\text{mask}}\}$ & $\{\mathcal{P}_{\text{mask}}\}$ \\
\bottomrule
\end{tabular}
}
\end{table}

This unified formulation ensures that a single generative model can seamlessly transition between full spatial projection and localized topological repair. Crucially, as outlined in Table~\ref{tab:multitask_configs}, this structure naturally accommodates our local geometric restoration task.
\subsection{Instruction-Tuning with Task-Aware Curriculum Schedule}
\label{sec:finetuning}

\begin{figure}
    \centering
    \includegraphics[width=\linewidth]{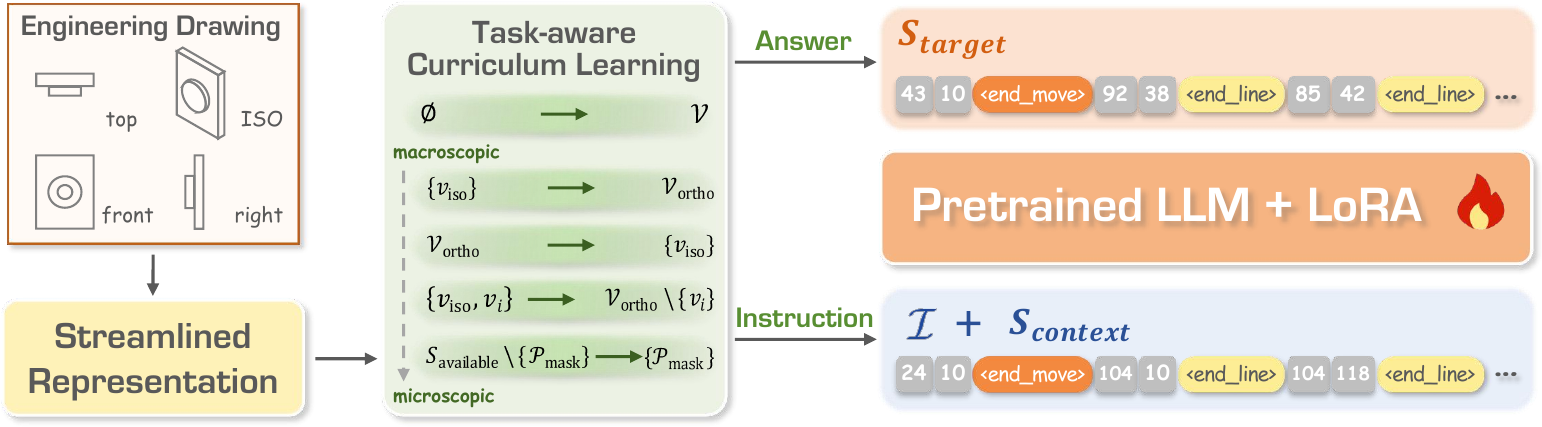}
\caption{\textbf{Overview of the proposed training framework.} We cast diverse multi-view generation sub-tasks into a unified sequence modeling paradigm, guided by a progressive microscopic-to-macroscopic optimization schedule. The task sampling distribution dynamically transitions from microscopic local topology learning (Stage I) to partial view alignment (Stage II), and ultimately to macroscopic global spatial translation (Stage III) across training epochs.}
\label{fig:pipeline}    \label{fig:pipeline}
\end{figure}

To empower the LLM with comprehensive multi-view spatial reasoning, we cast all generation sub-tasks detailed in Table~\ref{tab:multitask_configs} into a unified prediction framework, as illustrated in Figure \ref{fig:pipeline}. Rather than training specialized models, we employ a single pre-trained Llama-family~\cite{touvron2023llama} causal language model. To achieve this efficiently, we apply Low-Rank Adaptation (LoRA)~\cite{hu2022lora} alongside standard instruction-tuning practices. Specifically, we compute the cross-entropy loss $\mathcal{L}$ exclusively over the generative target $S_{\text{target}}$:
\begin{equation}
    \mathcal{L} = - \sum_{t=1}^{|S_{\text{target}}|} \log P(x_t \mid x_{<t}, \mathcal{I}, S_{\text{context}} ; \Theta) ,
\end{equation}
where $\Theta$ denotes the trainable LoRA parameters. By masking the instruction $\mathcal{I}$ and context $S_{\text{context}}$ during optimization, this unified objective allows the model to seamlessly alternate between diverse generation modes within the same training batch without gradient interference.

\paragraph{Microscopic-to-Macroscopic Task-Aware Curriculum.} 
Empirical evidence suggests that directly optimizing LLMs on complex, weakly-conditioned spatial generation tasks often leads to sub-optimal convergence and hallucinated topologies~\cite{bengio2009curriculum,wan2025wan}. To address this, we introduce a progressive, task-aware curriculum schedule that dynamically adjusts the sampling distribution of tasks across training epochs. 

Crucially, we treat \textit{Unconditional Generation} ($\emptyset \rightarrow \mathcal{V}$) as a continuous base task. Its sampling weight remains continuously active throughout the entire training process to robustly anchor the model's fundamental data distribution and generative capacity. Concurrently, the distribution of the remaining conditional tasks follows a smoothed optimization trajectory spanning three progressive stages:

\begin{itemize}
    \item \textbf{Stage I: Local Syntax and Topology (Early Epochs).} The conditional sampling distribution heavily favors microscopic structural tasks, specifically \textit{Path Level Semantic Masking}. This initial phase compels the model to establish robust coordinate syntax and localized geometric priors before tackling complex global spatial reasoning.
    \item \textbf{Stage II: Partial View Alignment (Middle Epochs).} As local geometric stability is acquired, we dynamically shift the conditional weights towards \textit{Partial View Completion} (e.g., inferring two missing orthographic views given the isometric and one available orthographic view). This stage forces the model to learn cross-view spatial alignment and correlation anchored by strong contextual cues.
    \item \textbf{Stage III: Global Spatial Translation (Late Epochs).} In the final stage, the distribution transitions predominantly to macroscopic spatial tasks, specifically \textit{Iso-to-Ortho} and \textit{Ortho-to-Iso} translations. Having progressively mastered local topology and partial alignment, the model is now optimized for rigorous 3D structural reasoning and deterministic projective geometry inference from highly compressed contexts.
\end{itemize}

\begin{figure}[htbp]
    \centering
    \includegraphics[width=\linewidth]{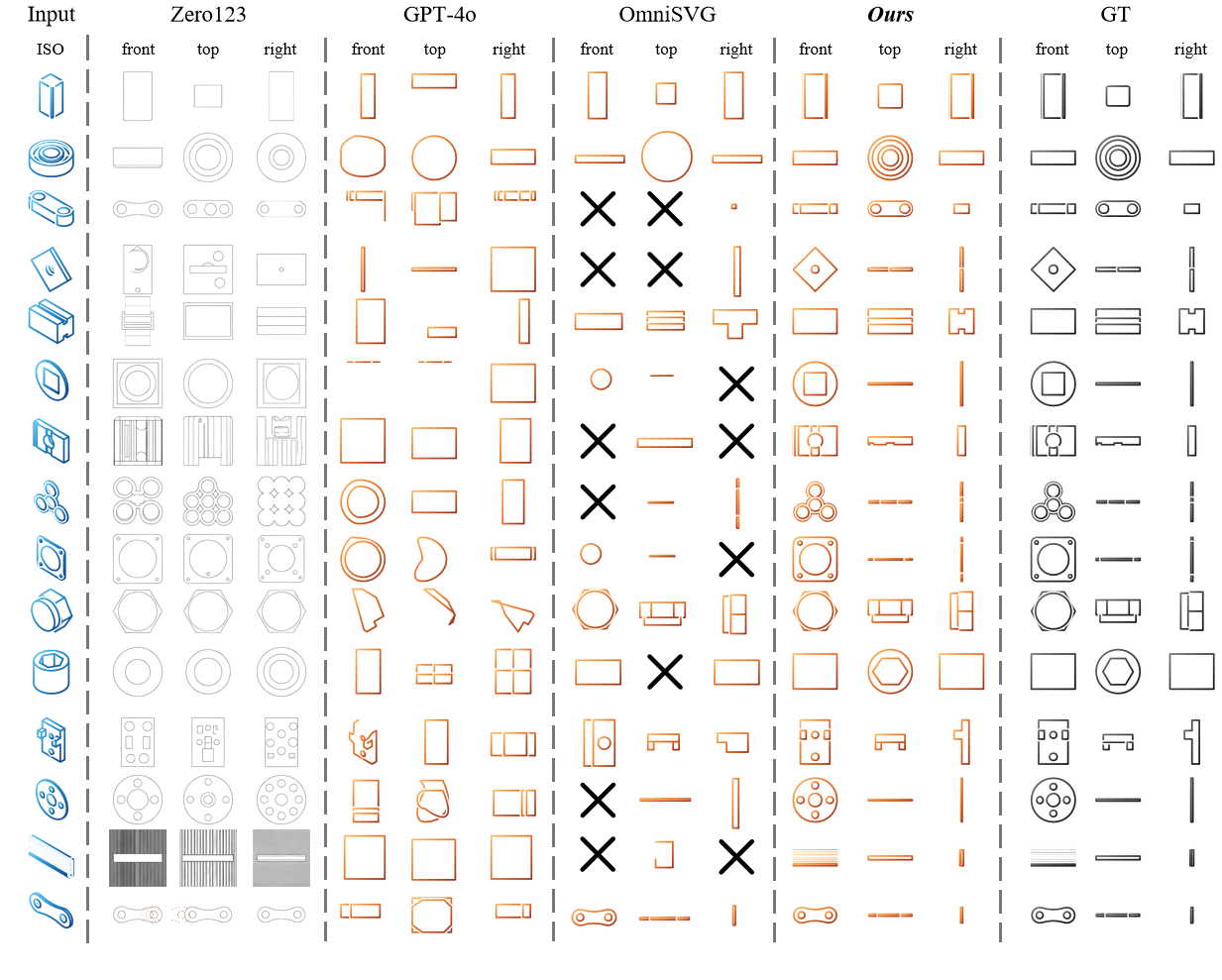} 
    \caption{Qualitative results on Isometric to Orthographic translation ($\{v_{\text{iso}}\} \rightarrow \mathcal{V}_{\text{ortho}}$).}
    \label{fig:iso2ortho}
\end{figure}

By adopting this microscopic-to-macroscopic optimization trajectory, our unified model sequentially builds its geometric reasoning capacity, seamlessly scaling from localized structural repair to unconstrained multi-view spatial imagination.

\section{Experiments} \label{sec: experiments}
\subsection{Experimental Setup} \label{sec: setup}
\paragraph{Implementation Details.}
Implementation Details. We instantiate DrawingsDreamer using LLaMA-3.2-3B~\cite{meta2024introducing} as the base LLM, chosen for its strong sequence modeling capabilities and competitive performance among open-source foundation models. To ensure parameter-efficient fine-tuning, we employ LoRA~\cite{hu2022lora} with a rank of $r=8$ and $\alpha=32$. Our dataset is constructed from DeepCAD~\cite{wu2021deepcad}, comprising 178,238 valid CAD models. We split the dataset at the CAD-model level into 160,414 training, 8,912 validation, and 8,912 test models, corresponding to a 90\%/5\%/5\% split. For the hardware setup, models with parameters up to 3B are trained on a cluster of eight NVIDIA RTX 4090 GPUs, while larger models are trained on eight NVIDIA L40 GPUs. We employ the AdamW optimizer~\cite{loshchilov2017decoupled} with a batch size of 16. We apply a cosine annealing learning rate schedule initialized at $5 \times 10^{-4}$ and train the model for 15 epochs.
\subsection{Metrics}
\label{sec:metrics}
To evaluate both the syntactic correctness and geometric fidelity of our generated outputs, we employ a comprehensive protocol tailored to distinct task categories.

\paragraph{Conditional Generation.} For translation and completion tasks (Iso-to-Ortho, Ortho-to-Iso, and Partial View Completion), we evaluate performance across three levels. Syntactically, we compute Command Accuracy ($\text{Acc}_{\text{cmd}}$) and Parameter Accuracy ($\text{Acc}_{\text{params}}$) to assess the exact match rate of structural tokens and the precision of numerical coordinates, respectively. Visually, we report PSNR, SSIM, and LPIPS on rasterized 2D multi-view SVGs. Geometrically, we evaluate cross-view geometric fidelity by uniformly sampling point clouds from the geometric paths of the generated SVGs across different views and computing the Chamfer Distance (CD) between them.

\paragraph{Unconditional Generation.} For this task, we measure the diversity and distributional alignment of the generated drawings against the dataset. Following standard shape generation protocols~\cite{wu2021deepcad}, we report Jensen-Shannon Divergence (JSD), Coverage (COV), and Maximum Mean Discrepancy (MMD).

\subsection{Performance Comparison with Existing Methods}
\label{sec:baselines}

\paragraph{Baselines.}
As direct baselines for this novel task are absent, we benchmark \textbf{DrawingsDreamer} against three representative paradigms: \textbf{GPT-4o (5-shot)} representing LLMs via in-context learning, \textbf{OmniSVG}~\cite{yang2025omnisvg} for recent SVG generation, and \textbf{Zero123}~\cite{liu2023zero} serving as a Iso-to-Ortho synthesis method.

\paragraph{Quantitative Results.}
The quantitative results demonstrate that our proposed unified model consistently achieves state-of-the-art performance across all metrics. \textbf{In terms of Cross-View Translation} (Table~\ref{tab:translation_results}), we evaluate the model's capabilities through $\{v_{\text{iso}}\} \rightarrow \mathcal{V}_{\text{ortho}}$ and $\mathcal{V}_{\text{ortho}} \rightarrow \{v_{\text{iso}}\}$ tasks. 
\begin{wraptable}{r}{0.55\textwidth}
  \centering
  \caption{Quantitative comparison on Unconditional Generation ($\emptyset \rightarrow \mathcal{V}$). JSD and MMD are multipled by $10^2$. \colorbox{red!20}{Red} and \colorbox{green!20}{green} cells denote the best and second-best results, respectively.}
  \label{tab:uncond_results}
  \begin{tabular}{l ccc}
    \toprule
    Method & JSD$\downarrow$ & COV$\uparrow$ & MMD$\downarrow$ \\
    \midrule
    GPT-4o (5-shot)& \cellcolor{green!20}1.17 & \cellcolor{green!20}33.12\% & \cellcolor{green!20}8.35 \\
    % DeepSVG & 2.51 & 27.6\% & 13.17 \\
    DrawingsDreamer & \cellcolor{red!20}0.61 & \cellcolor{red!20}66.24\% & \cellcolor{red!20}5.64 \\
    \bottomrule
  \end{tabular}
\end{wraptable}\textbf{DrawingsDreamer} significantly outperforms strong baselines, including GPT-4o, Zero123~\cite{liu2023zero}, and OmniSVG~\cite{yang2025omnisvg}. The performance gap, particularly with OmniSVG, can largely be attributed to its reliance on vision encoders pre-trained on natural images, which struggle to capture the rigorous spatial alignments and precise topological dependencies required for engineering drawings. Notably, our model maintains robust syntactic fidelity, achieving superior command and parameter accuracies across both translation directions. 

\textbf{For Partial View Completion} (Table~\ref{tab:completion_results}), which evaluates local geometric restoration, \textbf{DrawingsDreamer} excels in maintaining geometric fidelity and structural consistency within complex multi-view contexts. \textbf{Finally, regarding Unconditional Generation} (Table~\ref{tab:uncond_results}), our model achieves competitive JSD and MMD scores, demonstrating high generative diversity and distributional alignment. This confirms that our autoregressive paradigm effectively captures the underlying distribution of structured CAD blueprints, enabling the synthesis of complex engineering geometries from scratch.

\begin{table}
  \centering
  \caption{Quantitative comparison on Cross-View Translation tasks.  GPT-4o is enhanced with few-shot in-context learning. Specifically, each prompt comprises five exemplars randomly chosen from the training set. These exemplars include instructions and answer. \colorbox{red!20}{Red} and \colorbox{green!20}{green} cells denote the best and second-best results, respectively.}
  \label{tab:translation_results}
  \resizebox{\textwidth}{!}{
  \begin{tabular}{l cccccc cccccc}
    \toprule
    \multirow{2}{*}{Method} 
    & \multicolumn{6}{c}{$\{v_{\text{iso}}\} \rightarrow \mathcal{V}_{\text{ortho}}$} 
    & \multicolumn{6}{c}{$\mathcal{V}_{\text{ortho}} \rightarrow \{v_{\text{iso}}\}$} \\
    \cmidrule(lr){2-7} \cmidrule(lr){8-13}
    & \makecell{$Acc_{cmd}\uparrow$} & \makecell{$Acc_{params}\uparrow$} & PSNR$\uparrow$ & SSIM$\uparrow$ & LPIPS$\downarrow$ & CD$\downarrow$ 
    & \makecell{$Acc_{cmd}\uparrow$} & \makecell{$Acc_{params}\uparrow$} & PSNR$\uparrow$ & SSIM$\uparrow$ & LPIPS$\downarrow$ & CD$\downarrow$ \\
    \midrule
    GPT-4o(5shot) 
    & \cellcolor{green!20}46.44\% & \cellcolor{green!20}35.54\% & \cellcolor{green!20}33.53 & 0.80 & 0.19 & \cellcolor{green!20}0.34
    & \cellcolor{green!20}11.33\% & \cellcolor{green!20}56.68\% & \cellcolor{green!20}11.04 & \cellcolor{green!20}0.72 & \cellcolor{green!20}0.23 & \cellcolor{green!20}0.23 \\
    
    Zero123 
    & -- & -- & 24.93 & 0.85 & 0.16 & 0.75
    & -- & -- & -- & -- & -- & -- \\
    
    OmniSVG
    & -- & -- & 20.01 & \cellcolor{green!20}0.88 & \cellcolor{green!20}0.10 & 0.47
    & -- & -- & -- & -- & -- & -- \\
    
    DrawingsDreamer 
    & \cellcolor{red!20}89.77\% & \cellcolor{red!20}80.82\% & \cellcolor{red!20}50.16 & \cellcolor{red!20}0.94 & \cellcolor{red!20}0.03 & \cellcolor{red!20}0.02
    & \cellcolor{red!20}73.53\% & \cellcolor{red!20}68.36\% & \cellcolor{red!20}32.03 & \cellcolor{red!20}0.88 & \cellcolor{red!20}0.06 & \cellcolor{red!20}0.02 \\
    \bottomrule
  \end{tabular}
  }
\end{table}

\begin{table}[htbp]
    \centering
\caption{Ablation Study and Quantitative Results. The evaluation is grouped by task: Cross-View Translation (top) and Partial View Completion alongside Unconditional Generation (bottom). LPIPS, CD, JSD, and MMD are multiplied by $10^2$. \colorbox{red!20}{Red} and \colorbox{green!20}{green} cells denote the best and second-best results, respectively.}
\label{tab:abs}
    
    % --- Top Block: Cross-View Translation ---
    \vspace{0.1cm}
    \resizebox{\textwidth}{!}{
        \begin{tabular}{l cccccc cccccc}
            \toprule
            \multirow{2}{*}{Method} & 
            \multicolumn{6}{c}{$\{v_{\text{iso}}\} \rightarrow \mathcal{V}_{\text{ortho}}$} & 
            \multicolumn{6}{c}{$\{v_{\text{ortho}}\} \rightarrow \mathcal{V}_{\text{iso}}$} \\
            \cmidrule(lr){2-7} \cmidrule(lr){8-13}
            & $Acc_{cmd}\uparrow$ & $Acc_{params}\uparrow$ & PSNR$\uparrow$ & SSIM$\uparrow$ & LPIPS$\downarrow$ & CD$\downarrow$ 
            & $Acc_{cmd}{\uparrow}$ & $Acc_{params}\uparrow$ & PSNR$\uparrow$ & SSIM$\uparrow$ & LPIPS$\downarrow$ & CD$\downarrow$ \\
            \midrule
            w/o Path Mask & 
            86.35\% & 75.64\% & \cellcolor{green!20}48.99 & 0.91 & 3.53 & 3.17 & 
            73.18\% & 62.89\% & 31.77 & 0.84 & 6.51 & 2.85 \\
            Prefix Format & 
            88.11\% & \cellcolor{green!20}80.58\% & 47.12 & 0.87 & 4.01 & 2.83 & 
            67.11\% & 67.28\% & 30.60 & 0.86 & 6.67 & 2.74 \\
            Random Schedule & 
            87.49\% & 73.42\% & 40.91 & \cellcolor{green!20}0.92 & 4.52 & 3.03 & 
            70.91\% & \cellcolor{green!20}68.23\% & 31.64 & 0.83 & 6.43 & 2.59 \\
            Isolated: $\{v_{\text{iso}}\} \rightarrow \mathcal{V}_{\text{ortho}}$ & 
            \cellcolor{green!20}88.71\% & 79.22\% & 47.42 & \cellcolor{red!20}0.94 & \cellcolor{green!20}3.44 & \cellcolor{red!20}2.36 & 
            -- & -- & -- & -- & -- & -- \\
            Isolated: $\{v_{\text{ortho}}\} \rightarrow \mathcal{V}_{\text{iso}}$ & 
            -- & -- & -- & -- & -- & -- & 
            \cellcolor{red!20}75.58\% & 66.85\% & \cellcolor{red!20}33.68 & \cellcolor{red!20}0.89 & \cellcolor{green!20}5.58 & \cellcolor{green!20}2.45 \\
            Ours & 
            \cellcolor{red!20}89.77\%& \cellcolor{red!20}80.82\% & \cellcolor{red!20}50.16 & \cellcolor{red!20}0.94 & \cellcolor{red!20}3.30 & \cellcolor{green!20}2.41 & 
            \cellcolor{green!20}73.53\% & \cellcolor{red!20}68.36\% & \cellcolor{green!20}32.03 & \cellcolor{green!20}0.88 & \cellcolor{red!20}5.51 & \cellcolor{red!20}2.38 \\
            \bottomrule
        \end{tabular}
    }
    
    % --- Bottom Block: Completion & Unconditional ---
    \vspace{0.4cm} % Adds breathing room between the two blocks
    
    \resizebox{0.85\textwidth}{!}{
        \begin{tabular}{l cccccc ccc}
            \toprule
            \multirow{2}{*}{Method} & 
            \multicolumn{6}{c}{$\{v_{\text{iso}}, v_{\text{front}}\} \rightarrow \{v_{\text{top}}, v_{\text{right}}\}$} & 
            \multicolumn{3}{c}{$\emptyset \rightarrow \mathcal{V}$} \\
            \cmidrule(lr){2-7} \cmidrule(lr){8-10}
            & $Acc_{cmd}\uparrow$ & $Acc_{params}\uparrow$ & PSNR$\uparrow$ & SSIM$\uparrow$ & LPIPS$\downarrow$ & CD$\downarrow$ 
            & JSD$\downarrow$ & COV$\uparrow$ & MMD$\downarrow$ \\
            \midrule
            w/o Path Mask & 
            89.24\% & 79.91\% & 52.19 & 0.95 & \cellcolor{green!20}1.87 & 1.56 & 
            \cellcolor{green!20}0.63  & 60.77\% & 5.96 \\
            Prefix Format & 
            88.38\% & 80.07\% & \cellcolor{green!20}57.86 & 0.90 & 1.99 & 1.53 & 
            0.67  & \cellcolor{green!20}62.45\% & \cellcolor{green!20}5.70 \\
            Random Schedule & 
            \cellcolor{green!20}92.18\% & \cellcolor{red!20}86.39\% & 57.19 & \cellcolor{green!20}0.96 & 2.02 & \cellcolor{green!20}1.48 & 
            0.88  & 61.21\% & 5.85 \\
            Isolated: Partial & 
            85.02\% & \cellcolor{green!20}83.18\% & 55.52 & \cellcolor{green!20}0.96 & 2.31 & 1.73  & 
            -- & -- & -- \\
            Isolated: Uncond. & 
            -- & -- & -- & -- & -- & -- & 
            0.71 & 60.38\% & 5.79 \\
            Ours & 
            \cellcolor{red!20}92.82\% & 82.52\% & \cellcolor{red!20}58.61 & \cellcolor{red!20}0.97 & \cellcolor{red!20}1.83 & \cellcolor{red!20}1.41 & 
            \cellcolor{red!20}0.61 & \cellcolor{red!20}66.24\% & \cellcolor{red!20}5.64 \\
            \bottomrule
        \end{tabular}
    }
\end{table}

\paragraph{Qualitative Results.}

\begin{figure}[htbp]
    \centering
    \includegraphics[width=\linewidth]{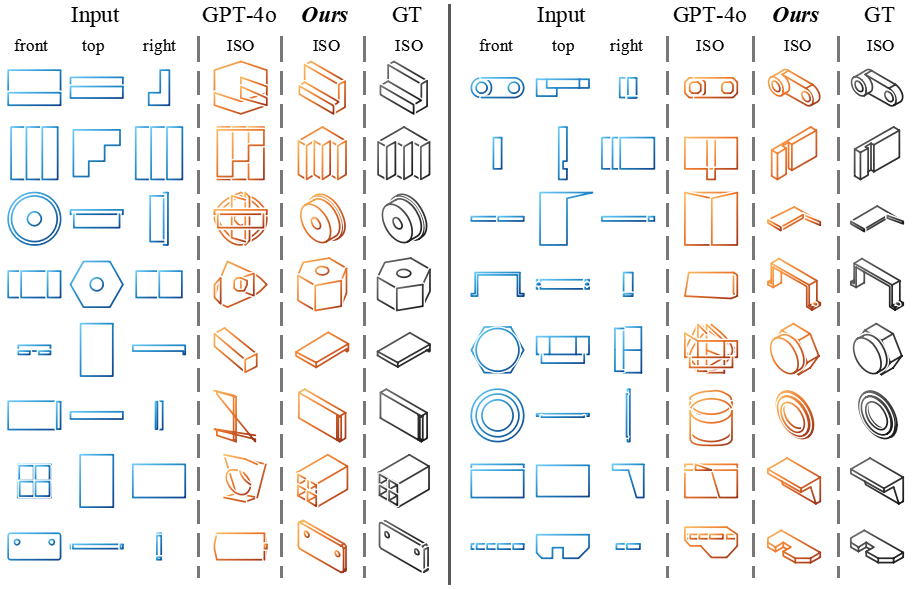}
    \caption{Qualitative results on Orthographic to Isometric translation ($\mathcal{V}_{\text{ortho}} \rightarrow \{v_{\text{iso}}\}$). }
    \label{fig:ortho2iso}
\end{figure}

\begin{figure}[htbp]
    \centering
    \includegraphics[width=\linewidth]{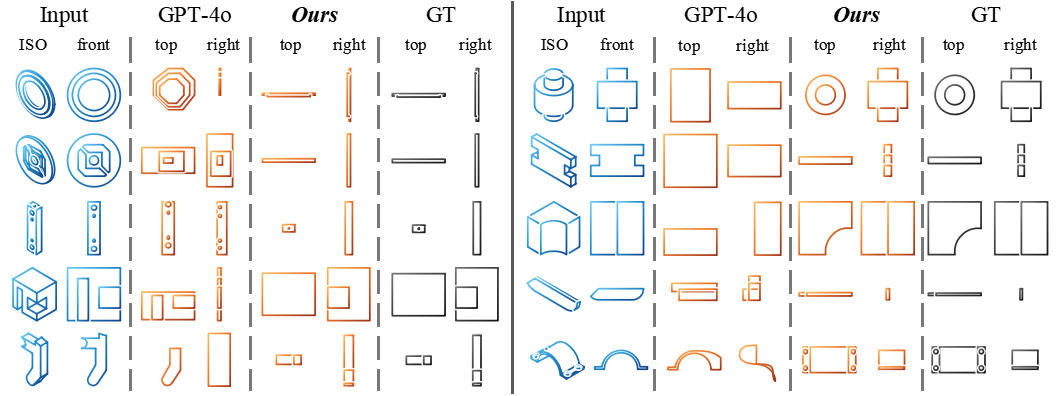}
    \caption{Qualitative results on Partial View Completion ($\{v_{\text{iso}}, v_{\text{front}}\} \rightarrow \{v_{\text{top}}, v_{\text{right}}\}$). Both the left and right panels display separate generation cases.}
    \label{fig:iso_front_to_top_right}
\end{figure}

Figure~\ref{fig:iso2ortho} illustrates the $\{v_{\text{iso}}\} \rightarrow \mathcal{V}_{\text{ortho}}$ translation, where \textbf{DrawingsDreamer} accurately reconstructs 3D geometries from isometric perspectives into aligned orthographic projections. Conversely, Figure~\ref{fig:ortho2iso} shows the inverse task ($\mathcal{V}_{\text{ortho}} \rightarrow \{v_{\text{iso}}\}$), demonstrating the model's ability to aggregate disjoint coordinates into a consistent isometric view. Furthermore, Figure~\ref{fig:iso_front_to_top_right} highlights our model's localized spatial reasoning on partial view completion. Even with complex internal contours, the generated views maintain strict geometric and structural alignment with the provided multi-view context.

\subsection{Ablation Studies}
\label{sec:ablations}

\begin{table}[htbp]
  \centering
  \caption{Quantitative evaluation on Partial View Completion ($\{v_{\text{iso}}, v_{\text{front}}\} \rightarrow \{v_{\text{top}}, v_{\text{right}}\}$). \colorbox{red!20}{Red} and \colorbox{green!20}{green} cells denote the best and second-best results}
  \label{tab:completion_results}
  \begin{tabular}{l cccccc}
    \toprule
    Method & \makecell{$Acc_{cmd}\uparrow$} & \makecell{$Acc_{params}\uparrow$} & PSNR$\uparrow$ & SSIM$\uparrow$ & LPIPS$\downarrow$ & CD$\downarrow$ \\
    \midrule
    GPT-4o (5-shot) & \cellcolor{green!20}65.48\% & \cellcolor{green!20}30.13\% & \cellcolor{green!20}16.52 & \cellcolor{green!20}0.83 & \cellcolor{green!20}0.16 & \cellcolor{green!20}0.30 \\
    DrawingsDreamer & \cellcolor{red!20}92.82\% & \cellcolor{red!20}82.52\% & \cellcolor{red!20}58.61 & \cellcolor{red!20}0.97 & \cellcolor{red!20}0.02 & \cellcolor{red!20}0.01 \\
    \bottomrule
  \end{tabular}
\end{table}

To rigorously evaluate our proposed components, we conduct comprehensive ablation studies as summarized in Table~\ref{tab:abs}.

\paragraph{Unified Training vs. Isolated Optimization.}
Compared to training specialized models for isolated tasks (e.g., $\{v_{\text{iso}}\} \rightarrow \mathcal{V}_{\text{ortho}}$), our unified multitask paradigm consistently achieves superior performance. This confirms that joint optimization effectively transfers knowledge across tasks to enhance overall multi-view spatial reasoning.
\paragraph{Task-Aware Curriculum Schedule.}
Replacing our progressive three-stage curriculum with random task sampling leads to sub-optimal convergence and degraded Chamfer Distance. This highlights the necessity of a structured transition, anchoring local syntax before scaling to global spatial translation, to stabilize the generative process.

\paragraph{Path Level Semantic Masking.}
Removing this microscopic structural task during training degrades geometric fidelity and command accuracy in the generated multiview sequences. Forcing the model to explicitly learn basic geometric structures serves as a crucial stepping stone for coherent macroscopic generation.

\paragraph{Postfix Tokenization vs. Prefix Formats.}
Replacing our hierarchical postfix tokenization with standard prefix representations (e.g., \texttt{M}, \texttt{L}, \texttt{C} sequencing)  drops parameter accuracy. Regressing coordinates before assigning semantic boundaries aligns better with the innate sequence modeling capabilities of causal LLMs in engineering drawing scenarios.

\section{Conclusion}

In this paper, we introduced \textbf{DrawingsDreamer}, a unified generative framework that frames multi-view engineering drawing synthesis as a pure sequence modeling task. By leveraging a streamlined representation with hierarchical postfix tokenization and a progressive task-aware curriculum, our approach effectively captures rigorous cross-view spatial alignments and complex geometric dependencies. Extensive evaluations demonstrate that \textbf{DrawingsDreamer} sets a new state of the art in generating geometrically precise parametric engineering drawings. We anticipate that expanding its geometric vocabulary to support more complex parametric constraints will further advance the integration of general-purpose sequence modeling into professional AI-assisted industrial design workflows.

\bibliographystyle{abbrv} 
\bibliography{references}

@inproceedings{qin2025drawing2cad,
  title={Drawing2CAD: Sequence-to-Sequence Learning for CAD Generation from Vector Drawings},
  author={Qin, Feiwei and Lu, Shichao and Hou, Junhao and Wang, Changmiao and Fang, Meie and Liu, Ligang},
  booktitle={ACM MM},

  year={2025}
}

@inproceedings{wu2021deepcad,
  title={Deepcad: A deep generative network for computer-aided design models},
  author={Wu, Rundi and Xiao, Chang and Zheng, Changxi},
  booktitle={ICCV},

  year={2021}
}

@article{xu2024brepgen,
  title={Brepgen: A b-rep generative diffusion model with structured latent geometry},
  author={Xu, Xiang and Lambourne, Joseph and Jayaraman, Pradeep and Wang, Zhengqing and Willis, Karl and Furukawa, Yasutaka},
  journal={TOG},

  year={2024},
  publisher={ACM New York, NY, USA}
}

@inproceedings{li2025dtgbrepgen,
  title={Dtgbrepgen: A novel b-rep generative model through decoupling topology and geometry},
  author={Li, Jing and Fu, Yihang and Chen, Falai},
  booktitle={CVPR},
  year={2025}
}

@article{liu2026hidigen,
  title={HiDiGen: Hierarchical Diffusion for B-Rep Generation with Explicit Topological Constraints},
  author={Liu, Shurui and Chen, Weide and Wu, Ancong},
  journal={arXiv preprint arXiv:2604.02847},
  year={2026}
}

@inproceedings{khan2024cad,
  title={Cad-signet: Cad language inference from point clouds using layer-wise sketch instance guided attention},
  author={Khan, Mohammad Sadil and Dupont, Elona and Ali, Sk Aziz and Cherenkova, Kseniya and Kacem, Anis and Aouada, Djamila},
  booktitle={CVPR},

  year={2024}
}

@article{meta2024introducing,
  title={Introducing meta llama 3: The most capable openly available llm to date},
  author={Meta, AI},
  journal={Meta AI},
  year={2024}
}

@article{hu2022lora,
  title={Lora: Low-rank adaptation of large language models.},
  author={Hu, Edward J and Shen, Yelong and Wallis, Phillip and Allen-Zhu, Zeyuan and Li, Yuanzhi and Wang, Shean and Wang, Liang and Chen, Weizhu and others},
  journal={ICLR},
  year={2022}
}

@inproceedings{loshchilov2017decoupled,
  author       = {Ilya Loshchilov and
                  Frank Hutter},
  title        = {Decoupled Weight Decay Regularization},
  booktitle    = {ICLR},
  year         = {2019}
}

@article{vaswani2017attention,
  title={Attention is all you need},
  author={Vaswani, Ashish and Shazeer, Noam and Parmar, Niki and Uszkoreit, Jakob and Jones, Llion and Gomez, Aidan N and Kaiser, {\L}ukasz and Polosukhin, Illia},
  journal={NeurIPS},

  year={2017}
}

@inproceedings{koch2019abc,
  title={Abc: A big cad model dataset for geometric deep learning},
  author={Koch, Sebastian and Matveev, Albert and Jiang, Zhongshi and Williams, Francis and Artemov, Alexey and Burnaev, Evgeny and Alexa, Marc and Zorin, Denis and Panozzo, Daniele},
  booktitle={CVPR},

  year={2019}
}

@article{willis2021fusion,
  title={Fusion 360 gallery: A dataset and environment for programmatic cad construction from human design sequences},
  author={Willis, Karl DD and Pu, Yewen and Luo, Jieliang and Chu, Hang and Du, Tao and Lambourne, Joseph G and Solar-Lezama, Armando and Matusik, Wojciech},
  journal={TOG},
  year={2021},
  publisher={ACM New York, NY, USA}
}

@inproceedings{rukhovich2025cad,
  title={Cad-recode: Reverse engineering cad code from point clouds},
  author={Rukhovich, Danila and Dupont, Elona and Mallis, Dimitrios and Cherenkova, Kseniya and Kacem, Anis and Aouada, Djamila},
  booktitle={ICCV},

  year={2025}
}

@inproceedings{wang2025cadgpt,
  title={CAD-GPT: Synthesising CAD construction sequence with spatial reasoning-enhanced multimodal LLMs},
  author={Wang, Siyu and Chen, Cailian and Le, Xinyi and Xu, Qimin and Xu, Lei and Zhang, Yanzhou and Yang, Jie},
  booktitle={AAAI},

  year={2025}
}

@inproceedings{alrashedy2024generating,
author       = {Kamel Alrashedy and
                  Pradyumna Tambwekar and
                  Zulfiqar Haider Zaidi and
                  Megan Langwasser and
                  Wei Xu and
                  Matthew C. Gombolay},
  title        = {Generating {CAD} Code with Vision-Language Models for 3D Designs},
  booktitle    = {ICLR},
  year         = {2025},
}

@article{yavartanoo2024text2cad,
  title={Text2CAD: Text to 3D CAD generation via technical drawings},
  author={Yavartanoo, Mohsen and Hong, Sangmin and Neshatavar, Reyhaneh and Lee, Kyoung Mu},
  journal={arXiv preprint arXiv:2411.06206},
  year={2024}
}

@inproceedings{chen2025cadcrafter,
  title={Cadcrafter: Generating computer-aided design models from unconstrained images},
  author={Chen, Cheng and Wei, Jiacheng and Chen, Tianrun and Zhang, Chi and Yang, Xiaofeng and Zhang, Shangzhan and Yang, Bingchen and Foo, Chuan-Sheng and Lin, Guosheng and Huang, Qixing and others},
  booktitle={CVPR},

  year={2025}
}

@inproceedings{li2025caddreamer,
  title={CADDreamer: CAD object generation from single-view images},
  author={Li, Yuan and Lin, Cheng and Liu, Yuan and Long, Xiaoxiao and Zhang, Chenxu and Wang, Ningna and Li, Xin and Wang, Wenping and Guo, Xiaohu},
  booktitle={CVPR},

  year={2025}
}

@article{khan2026dreamcad,
  title={DreamCAD: Scaling Multi-modal CAD Generation using Differentiable Parametric Surfaces},
  author={Khan, Mohammad Sadil and Usama, Muhammad and Potamias, Rolandos Alexandros and Stricker, Didier and Afzal, Muhammad Zeshan and Deng, Jiankang and Elezi, Ismail},
  journal={arXiv preprint arXiv:2603.05607},
  year={2026}
}

@article{gong2006reconstruction111,
  title={Reconstruction of 3D curvilinear wire-frame from three orthographic views},
  author={Gong, Jie-Hui and Zhang, Gui-Fang and Zhang, Hui and Sun, Jia-Guang},
  journal={Computers \& Graphics},

  year={2006},
  publisher={Elsevier}
}

@article{kuo1998reconstruction222,
  title={Reconstruction of quadric surface solids from three-view engineering drawings},
  author={Kuo, Mu-Hsing},
  journal={Computer-Aided Design},

  year={1998},
  publisher={Elsevier}
}

@article{liu2001reconstruction333,
  title={Reconstruction of curved solids from engineering drawings},
  author={Liu, Shi-Xia and Hu, Shi-Min and Chen, Yu-Jian and Sun, Jia-Guang},
  journal={Computer-Aided Design},

  year={2001},
  publisher={Elsevier}
}

@inproceedings{wang1993survey444,
  title={A survey of 3D solid reconstruction from 2D projection line drawings},
  author={Wang, Weidong and Grinstein, Georges G},
  booktitle={Computer Graphics Forum},

  year={1993},
  organization={Wiley Online Library}
}

@article{furferi20102d555,
  title={From 2D orthographic views to 3D pseudo-wireframe: An automatic procedure},
  author={Furferi, Rocco and Governi, Lapo and Palai, Matteo and Volpe, Yary and others},
  journal={International Journal of Computer Applications},

  year={2010}
}

@article{camba2022sketch,
  title={Sketch-based modeling in mechanical engineering design: Current status and opportunities},
  author={Camba, Jorge D and Company, Pedro and Naya, Ferran},
  journal={Computer-Aided Design},

  year={2022},
  publisher={Elsevier}
}

@article{harish2021photo2cad,
  title={Photo2CAD: Automated 3D solid reconstruction from 2D drawings using OpenCV},
  author={Harish, Ajay B and Prasad, Abhishek Rajendra},
  journal={arXiv preprint arXiv:2101.04248},
  year={2021}
}

@article{puhachov2023reconstruction,
  title={Reconstruction of machine-made shapes from bitmap sketches},
  author={Puhachov, Ivan and Martens, Cedric and Kry, Paul G and Bessmeltsev, Mikhail},
  journal={TOG},

  year={2023},
  publisher={ACM New York, NY, USA}
}

@inproceedings{wang20252d,
  title={From 2d cad drawings to 3d parametric models: A vision-language approach},
  author={Wang, Xilin and Zheng, Jia and Hu, Yuanchao and Zhu, Hao and Yu, Qian and Zhou, Zihan},
  booktitle={AAAI},
  year={2025}
}

@article{zhang2023automatic,
  title={Automatic 3D CAD models reconstruction from 2D orthographic drawings},
  author={Zhang, Chao and Pinqui{\'e}, Romain and Polette, Arnaud and Carasi, Gregorio and De Charnace, Henri and Pernot, Jean-Philippe},
  journal={Computers \& Graphics},
  year={2023},
  publisher={Elsevier}
}

@article{touvron2023llama,
  title={Llama: Open and efficient foundation language models},
  author={Touvron, Hugo and Lavril, Thibaut and Izacard, Gautier and Martinet, Xavier and Lachaux, Marie-Anne and Lacroix, Timoth{\'e}e and Rozi{\`e}re, Baptiste and Goyal, Naman and Hambro, Eric and Azhar, Faisal and others},
  journal={arXiv preprint arXiv:2302.13971},
  year={2023}
}

@article{achiam2023gpt,
  title={Gpt-4 technical report},
  author={Achiam, Josh and Adler, Steven and Agarwal, Sandhini and Ahmad, Lama and Akkaya, Ilge and Aleman, Florencia Leoni and Almeida, Diogo and Altenschmidt, Janko and Altman, Sam and Anadkat, Shyamal and others},
  journal={arXiv preprint arXiv:2303.08774},
  year={2023}
}

@article{yang2025qwen3,
  title={Qwen3 technical report},
  author={Yang, An and Li, Anfeng and Yang, Baosong and Zhang, Beichen and Hui, Binyuan and Zheng, Bo and Yu, Bowen and Gao, Chang and Huang, Chengen and Lv, Chenxu and others},
  journal={arXiv preprint arXiv:2505.09388},
  year={2025}
}

@article{das2025security,
  title={Security and privacy challenges of large language models: A survey},
  author={Das, Badhan Chandra and Amini, M Hadi and Wu, Yanzhao},
  journal={ACM Computing Surveys},
  year={2025},
  publisher={ACM New York, NY}
}

@inproceedings{li2023fine,
author       = {Juncheng Li and
                  Kaihang Pan and
                  Zhiqi Ge and
                  Minghe Gao and
                  Wei Ji and
                  Wenqiao Zhang and
                  Tat{-}Seng Chua and
                  Siliang Tang and
                  Hanwang Zhang and
                  Yueting Zhuang},
  title        = {Fine-tuning Multimodal LLMs to Follow Zero-shot Demonstrative Instructions},
  booktitle    = {ICLR},
  year         = {2024},

}

@article{zhang2025collm,
  title={Collm: Integrating collaborative embeddings into large language models for recommendation},
  author={Zhang, Yang and Feng, Fuli and Zhang, Jizhi and Bao, Keqin and Wang, Qifan and He, Xiangnan},
  journal={TKDE},
  year={2025},
  publisher={IEEE}
}

@inproceedings{shojaee2024llm,
  author       = {Parshin Shojaee and
                  Kazem Meidani and
                  Shashank Gupta and
                  Amir Barati Farimani and
                  Chandan K. Reddy},
  title        = {{LLM-SR:} Scientific Equation Discovery via Programming with Large
                  Language Models},
  booktitle    = {ICLR},
  year         = {2025}
}

@inproceedings{zhang2024llama,
  title={LLaMA-adapter: Efficient fine-tuning of large language models with zero-initialized attention},
  author={Zhang, Renrui and Han, Jiaming and Liu, Chris and Zhou, Aojun and Lu, Pan and Qiao, Yu and Li, Hongsheng and Gao, Peng},
  booktitle={ICLR},
  year={2024}
}

@article{jiang2023motiongpt,
  title={Motiongpt: Human motion as a foreign language},
  author={Jiang, Biao and Chen, Xin and Liu, Wen and Yu, Jingyi and Yu, Gang and Chen, Tao},
  journal={NeurIPS},
  year={2023}
}

@article{gu2025effectiveness,
  title={On the effectiveness of large language models in domain-specific code generation},
  author={Gu, Xiaodong and Chen, Meng and Lin, Yalan and Hu, Yuhan and Zhang, Hongyu and Wan, Chengcheng and Wei, Zhao and Xu, Yong and Wang, Juhong},
  journal={ACM Transactions on Software Engineering and Methodology},
  year={2025},
  publisher={ACM New York, NY}
}

@article{dong2025codescore,
  title={Codescore: Evaluating code generation by learning code execution},
  author={Dong, Yihong and Ding, Jiazheng and Jiang, Xue and Li, Ge and Li, Zhuo and Jin, Zhi},
  journal={ACM Transactions on Software Engineering and Methodology},
  year={2025},
  publisher={ACM New York, NY}
}

@article{li2025brepgpt,
  title={BrepGPT: Autoregressive B-rep Generation with Voronoi Half-Patch},
  author={Li, Pu and Zhang, Wenhao and Quan, Weize and Zhang, Biao and Wonka, Peter and Yan, Dongming},
  journal={TOG},
  year={2025},
  publisher={ACM New York, NY, USA}
}

@inproceedings{xu2025autobrep,
  title={AutoBrep: Autoregressive B-Rep Generation with Unified Topology and Geometry},
  author={Xu, Xiang and Jayaraman, Pradeep and Lambourne, Joseph and Liu, Yilin and Malpure, Durvesh and Meltzer, Pete},
  booktitle={SIGGRAPH},
  year={2025}
}

@inproceedings{li2025stitch,
  title={Stitch-A-Shape: Bottom-up Learning for B-Rep Generation},
  author={Li, Pu and Zhang, Wenhao and Chen, Jinglu and Yan, Dongming},
  booktitle={SIGGRAPH},
  year={2025}
}

@inproceedings{lee2025brepdiff,
  title={Brepdiff: Single-stage b-rep diffusion model},
  author={Lee, Mingi and Zhang, Dongsu and Jambon, Cl{\'e}ment and Kim, Young Min},
  booktitle={SIGGRAPH},
  year={2025}
}

@inproceedings{wu2024cadvlm,
  title={Cadvlm: Bridging language and vision in the generation of parametric cad sketches},
  author={Wu, Sifan and Khasahmadi, Amir Hosein and Katz, Mor and Jayaraman, Pradeep Kumar and Pu, Yewen and Willis, Karl and Liu, Bang},
  booktitle={ECCV},
  year={2024},
  organization={Springer}
}

@inproceedings{li2025revisiting,
  title={Revisiting cad model generation by learning raster sketch},
  author={Li, Pu and Zhang, Wenhao and Guo, Jianwei and Chen, Jinglu and Yan, Dong-Ming},
  booktitle={AAAI},
  year={2025}
}

@article{clouatre2019figr,
  title={Figr: Few-shot image generation with reptile},
  author={Clou{\^a}tre, Louis and Demers, Marc},
  journal={arXiv preprint arXiv:1901.02199},
  year={2019}
}

@article{kocetkov2022stack,
author       = {Denis Kocetkov and
                  Raymond Li and
                  Loubna Ben Allal and
                  Jia Li and
                  Chenghao Mou and
                  Yacine Jernite and
                  Margaret Mitchell and
                  Carlos Mu{\~{n}}oz Ferrandis and
                  Sean Hughes and
                  Thomas Wolf and
                  Dzmitry Bahdanau and
                  Leandro von Werra and
                  Harm de Vries},
  title        = {The Stack: 3 {TB} of permissively licensed source code},
  journal      = {Trans. Mach. Learn. Res.},

  year         = {2023}
 
}

@article{wang2021deepvecfont,
  title={Deepvecfont: synthesizing high-quality vector fonts via dual-modality learning},
  author={Wang, Yizhi and Lian, Zhouhui},
  journal={TOG},
  year={2021},
  publisher={ACM New York, NY, USA}
}

@inproceedings{yang2025omnisvg,
 author = {Yang, Yiying and Cheng, Wei and Chen, Sijin and Zeng, Xianfang and Yin, Fukun and Zhang, Jiaxu and Wang, Liao and Yu, Gang and Ma, Xingjun and Jiang, Yu-Gang},
 booktitle = {NeurIPS},
 publisher = {Curran Associates, Inc.},
 title = {OmniSVG:  A Unified Scalable Vector Graphics Generation Model},
 year = {2025}
}

@inproceedings{ha2017neural,
  author       = {David Ha and
                  Douglas Eck},
  title        = {A Neural Representation of Sketch Drawings},
  booktitle    = {ICLR},
  year         = {2018},
}

@inproceedings{reddy2021im2vec,
  title={Im2vec: Synthesizing vector graphics without vector supervision},
  author={Reddy, Pradyumna and Gharbi, Michael and Lukac, Michal and Mitra, Niloy J},
  booktitle={CVPR},
  year={2021}
}

@inproceedings{hu2024supersvg,
  title={Supersvg: Superpixel-based scalable vector graphics synthesis},
  author={Hu, Teng and Yi, Ran and Qian, Baihong and Zhang, Jiangning and Rosin, Paul L and Lai, Yu-Kun},
  booktitle={CVPR},
  year={2024}
}

@inproceedings{song2023clipvg,
  title={Clipvg: Text-guided image manipulation using differentiable vector graphics},
  author={Song, Yiren and Shao, Xuning and Chen, Kang and Zhang, Weidong and Jing, Zhongliang and Li, Minzhe},
  booktitle={AAAI},
  year={2023}
}

@inproceedings{ma2022towards,
  title={Towards layer-wise image vectorization},
  author={Ma, Xu and Zhou, Yuqian and Xu, Xingqian and Sun, Bin and Filev, Valerii and Orlov, Nikita and Fu, Yun and Shi, Humphrey},
  booktitle={CVPR},
  year={2022}
}

@article{li2020differentiable,
  title={Differentiable vector graphics rasterization for editing and learning},
  author={Li, Tzu-Mao and Luk{\'a}{\v{c}}, Michal and Gharbi, Micha{\"e}l and Ragan-Kelley, Jonathan},
  journal={TOG},
  year={2020},
}

@article{carlier2020deepsvg,
  title={Deepsvg: A hierarchical generative network for vector graphics animation},
  author={Carlier, Alexandre and Danelljan, Martin and Alahi, Alexandre and Timofte, Radu},
  journal={Advances in Neural Information Processing Systems},
  year={2020}
}

@inproceedings{lopes2019learned,
  title={A learned representation for scalable vector graphics},
  author={Lopes, Raphael Gontijo and Ha, David and Eck, Douglas and Shlens, Jonathon},
  booktitle={ICCV},
  year={2019}
}

@inproceedings{tang2024strokenuwa,
  author       = {Zecheng Tang and
                  Chenfei Wu and
                  Zekai Zhang and
                  Minheng Ni and
                  Shengming Yin and
                  Yu Liu and
                  Zhengyuan Yang and
                  Lijuan Wang and
                  Zicheng Liu and
                  Juntao Li and
                  Nan Duan},
  title        = {StrokeNUWA - Tokenizing Strokes for Vector Graphic Synthesis},
  booktitle    = {ICML},
  year         = {2024},
}

@article{su2023marvel,
  title={Marvel: Raster gray-level manga vectorization via primitive-wise deep reinforcement learning},
  author={Su, Hao and Liu, Xuefeng and Niu, Jianwei and Cui, Jiahe and Wan, Ji and Wu, Xinghao and Wang, Nana},
  journal={TCSVT},
  year={2023},
}

@inproceedings{tian2022modern,
  title={Modern evolution strategies for creativity: Fitting concrete images and abstract concepts},
  author={Tian, Yingtao and Ha, David},
  booktitle={EvoMUSART},
  year={2022}
}

@inproceedings{wu2025chat2svg,
  title={Chat2svg: Vector graphics generation with large language models and image diffusion models},
  author={Wu, Ronghuan and Su, Wanchao and Liao, Jing},
  booktitle={CVPR},
  year={2025}
}

@inproceedings{rodriguez2025starvector,
  title={Starvector: Generating scalable vector graphics code from images and text},
  author={Rodriguez, Juan A and Puri, Abhay and Agarwal, Shubham and Laradji, Issam H and Rodriguez, Pau and Rajeswar, Sai and Vazquez, David and Pal, Christopher and Pedersoli, Marco},
  booktitle={CVPR},
  year={2025}
}

@inproceedings{xing2025empowering,
  title={Empowering llms to understand and generate complex vector graphics},
  author={Xing, Ximing and Hu, Juncheng and Liang, Guotao and Zhang, Jing and Xu, Dong and Yu, Qian},
  booktitle={CVPR},
  year={2025}
}

@inproceedings{bengio2009curriculum,
  title={Curriculum learning},
  author={Bengio, Yoshua and Louradour, J{\'e}r{\^o}me and Collobert, Ronan and Weston, Jason},
  booktitle={ICML},
  year={2009}
}

@article{wan2025wan,
  title={Wan: Open and advanced large-scale video generative models},
  author={Wan, Team and Wang, Ang and Ai, Baole and Wen, Bin and Mao, Chaojie and Xie, Chen-Wei and Chen, Di and Yu, Feiwu and Zhao, Haiming and Yang, Jianxiao and others},
  journal={arXiv preprint arXiv:2503.20314},
  year={2025}
}

@inproceedings{liu2023zero,
  title={Zero-1-to-3: Zero-shot one image to 3d object},
  author={Liu, Ruoshi and Wu, Rundi and Van Hoorick, Basile and Tokmakov, Pavel and Zakharov, Sergey and Vondrick, Carl},
  booktitle={Proceedings of the IEEE/CVF international conference on computer vision},
  pages={9298--9309},
  year={2023}
}

%%%%%%%%%%%%%%%%%%%%%%%%%%%%%%%%%%%%%%%%%%%%%%%%%%%%%%%%%%%%

\appendix
\newpage
\appendix
\setcounter{figure}{0}
\renewcommand{\thefigure}{A\arabic{figure}}
{\huge{Appendix}}

Due to space limitations in the main paper, we provide additional results and discussions in this appendix, organized as follows:
\begin{itemize}
    \item Sec.~\ref{sec: dp} Data Processing Details

    \item Sec. ~\ref{sec: template} Prompting Templates and Task Configurations
    \item Sec. ~\ref{sec: scale} The Impact of Model Capacity
    \item Sec. ~\ref{sec: eval} Evaluation Metric Details

    \item Sec. ~\ref{sec: add} Additional Qualitative Results
    % \item Sec. ~\ref{sec: limit} Failure Cases, Limitations and Future Work
    \item Sec. ~\ref{sec: limit} Limitations and Future Work
\end{itemize}

\section{Data Processing Details}
\label{sec: dp}
Given a CAD program $\mathcal{J}$, we first reconstruct its corresponding
three-dimensional solid $\mathcal{B}$. We then generate a set of fixed-view
two-dimensional drawings from $\mathcal{B}$. The view set consists of three
orthographic views, namely Front, Right, and Top, together with one isometric
view denoted as Iso. Samples with missing or empty views are discarded.

To obtain geometrically stable multi-view inputs, all views are normalized into
a common $L \times L$ canvas, where $L=1000$. The three orthographic views share
a single scale factor, computed from their maximum spatial extent, so that their
relative sizes remain consistent across views. The isometric view is normalized
independently because its projected extent differs from the orthographic
projections.

Before tokenization, we canonicalize the SVG paths in each view. Whenever
closed contours can be recovered, we group path segments into loops and sort the
loops according to their spatial positions. For views with open or more complex
path structures, we use a deterministic traversal order while preserving all
visible path segments. This reduces the ambiguity introduced by arbitrary SVG
export order.

Each canonicalized path is then serialized into tokens. We retain three SVG
drawing commands: move, line, and cubic B\'ezier curve, corresponding to
$M$, $L$, and $C$, respectively. For a continuous coordinate
$x \in [0,L]$, its absolute quantized value is computed as

\begin{equation}
    q(x)=\mathrm{clip}\left(
\left\lfloor \frac{Qx}{L} \right\rfloor, 0, Q
\right),
\end{equation}

where $Q=256$ is the coordinate quantization resolution. The four view token
sequences are concatenated in the order Front, Right, Top, and Iso, with a
view-specific ending token inserted after each view. The resulting sequence
$\mathcal{S}$ is stored together with the model identifier and quantization
resolution as one training instance.

\begin{algorithm}[t]
\SetAlgoNlRelativeSize{-1}
\KwData{CAD model description $\mathcal{J}$}
\KwResult{Tokenized multi-view instance $(\mathrm{id}, Q, \mathcal{S})$}

$Q \leftarrow 256,\quad L \leftarrow 1000$ \tcp*{\textnormal{Quantization resolution and normalized canvas size}}

$\mathcal{B} \leftarrow \textit{reconstruct\_solid}(\mathcal{J})$ \tcp*{\textnormal{Recover the boundary representation from the CAD program}}

$\{\mathcal{R}_v\}_{v \in \mathcal{V}} \leftarrow \textit{render\_views}(\mathcal{B})$ \tcp*{\textnormal{Render standard 2D SVG projections}}

\If{$\neg\,\textit{valid\_views}(\{\mathcal{R}_v\})$}{
    \Return $\varnothing$ \tcp*{\textnormal{Discard models with missing or empty views}}
}

$\mathcal{V}_{o} \leftarrow \{\mathrm{Front}, \mathrm{Right}, \mathrm{Top}\}$,\quad
$\mathcal{V} \leftarrow \mathcal{V}_{o} \cup \{\mathrm{Iso}\}$

$s_o \leftarrow \textit{shared\_ortho\_scale}(\{\mathcal{R}_v\}_{v \in \mathcal{V}_{o}})$ 
\tcp*{\textnormal{Use a common scale for orthographic views}}

\ForEach{$v \in \mathcal{V}$}{
    $\mathcal{P}_v \leftarrow \textit{parse\_svg\_paths}(\mathcal{R}_v)$ 
    \tcp*{\textnormal{Extract path commands and continuous coordinates}}
    
    \eIf{$v \in \mathcal{V}_{o}$}{
        $\widetilde{\mathcal{P}}_v \leftarrow \textit{normalize\_view}(\mathcal{P}_v, s_o, L)$
        \tcp*{\textnormal{Normalize orthographic views with the shared scale}}
    }{
        $\widetilde{\mathcal{P}}_v \leftarrow \textit{normalize\_view}(\mathcal{P}_v, \textit{independent\_scale}(\mathcal{P}_v), L)$
        \tcp*{\textnormal{Normalize the isometric view independently}}
    }
    
    $\widehat{\mathcal{P}}_v \leftarrow \textit{order\_loops}(\widetilde{\mathcal{P}}_v)$
    \tcp*{\textnormal{Recover closed loops when possible and impose a canonical order}}
    
    $\mathcal{S}_v \leftarrow \textit{quantize\_and\_serialize}(\widehat{\mathcal{P}}_v, Q, L)$
    \tcp*{\textnormal{Convert SVG commands into discrete absolute-coordinate tokens}}
}
$\mathcal{S} \leftarrow
\mathcal{S}_{\mathrm{Front}} \Vert \langle\mathrm{end\_front}\rangle
\Vert \mathcal{S}_{\mathrm{Right}} \Vert \langle\mathrm{end\_right}\rangle
\Vert \mathcal{S}_{\mathrm{Top}} \Vert \langle\mathrm{end\_top}\rangle
\Vert \mathcal{S}_{\mathrm{Iso}} \Vert \langle\mathrm{end\_iso}\rangle$

\Return $(\mathrm{id}(\mathcal{J}), Q, \mathcal{S})$

\caption{\textit{Construction of multi-view SVG token sequences from a CAD model.}}
\label{alg:svg_data_generation}
\end{algorithm}

\section{Prompting Templates and Task Configurations}
\label{sec: template}

\begin{figure}[htbp]
    \centering
    \includegraphics[width=0.9\linewidth]{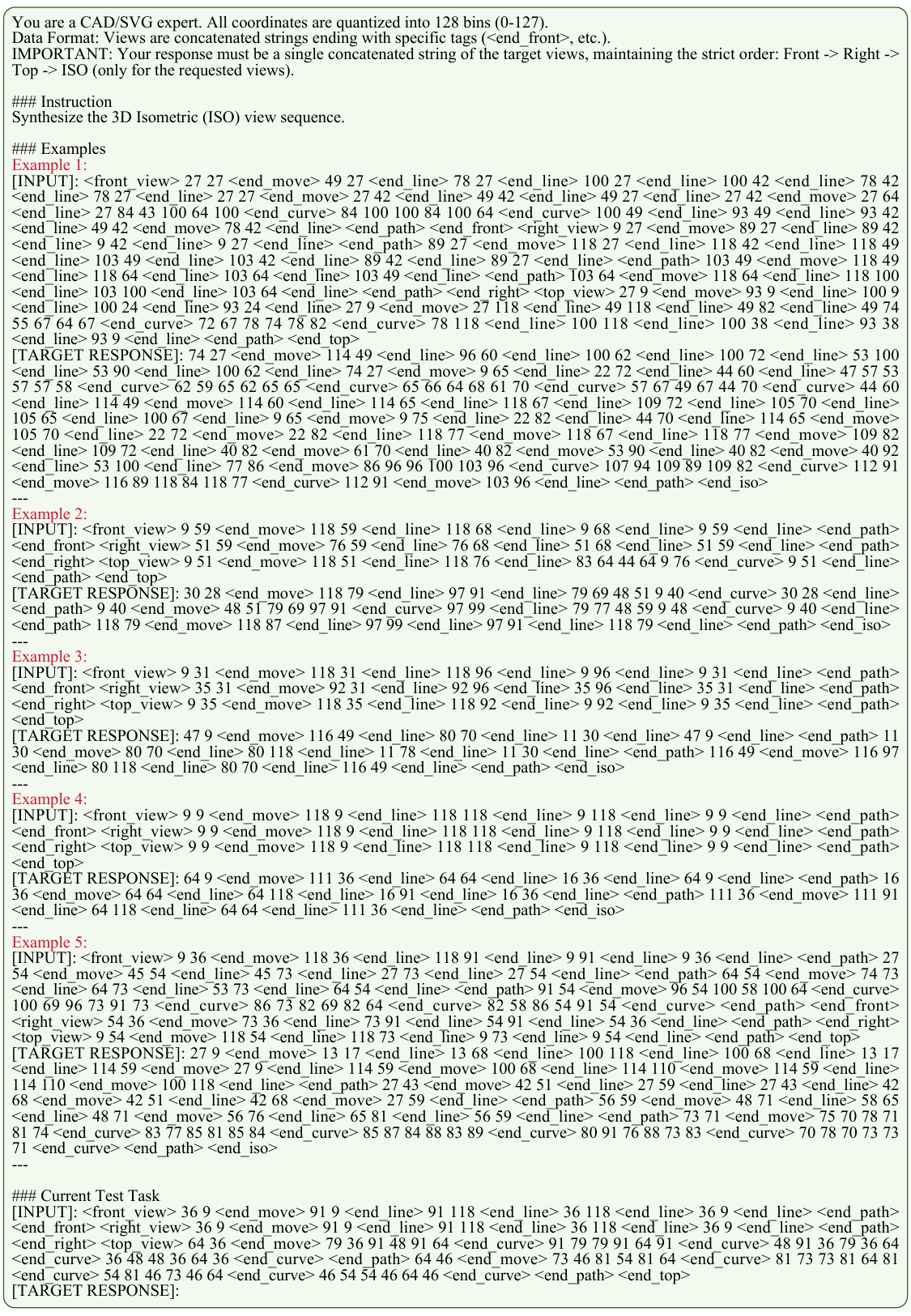}
    \caption{The 5-shot in-context learning prompt template used for the GPT-4o baseline evaluation.}
    \label{fig:gpt_template}
\end{figure}

As described in Section \ref{sec:baselines}, our evaluation of GPT-4o employs a 5-shot in-context learning setup. Figure~\ref{fig:gpt_template} illustrates the detailed prompt structure used in our experiments, which includes the system instruction and the layout of the five randomly sampled input-output exemplars prepended to the target query.

\begin{figure}[htbp]
    \centering
    \includegraphics[width=0.9\linewidth]{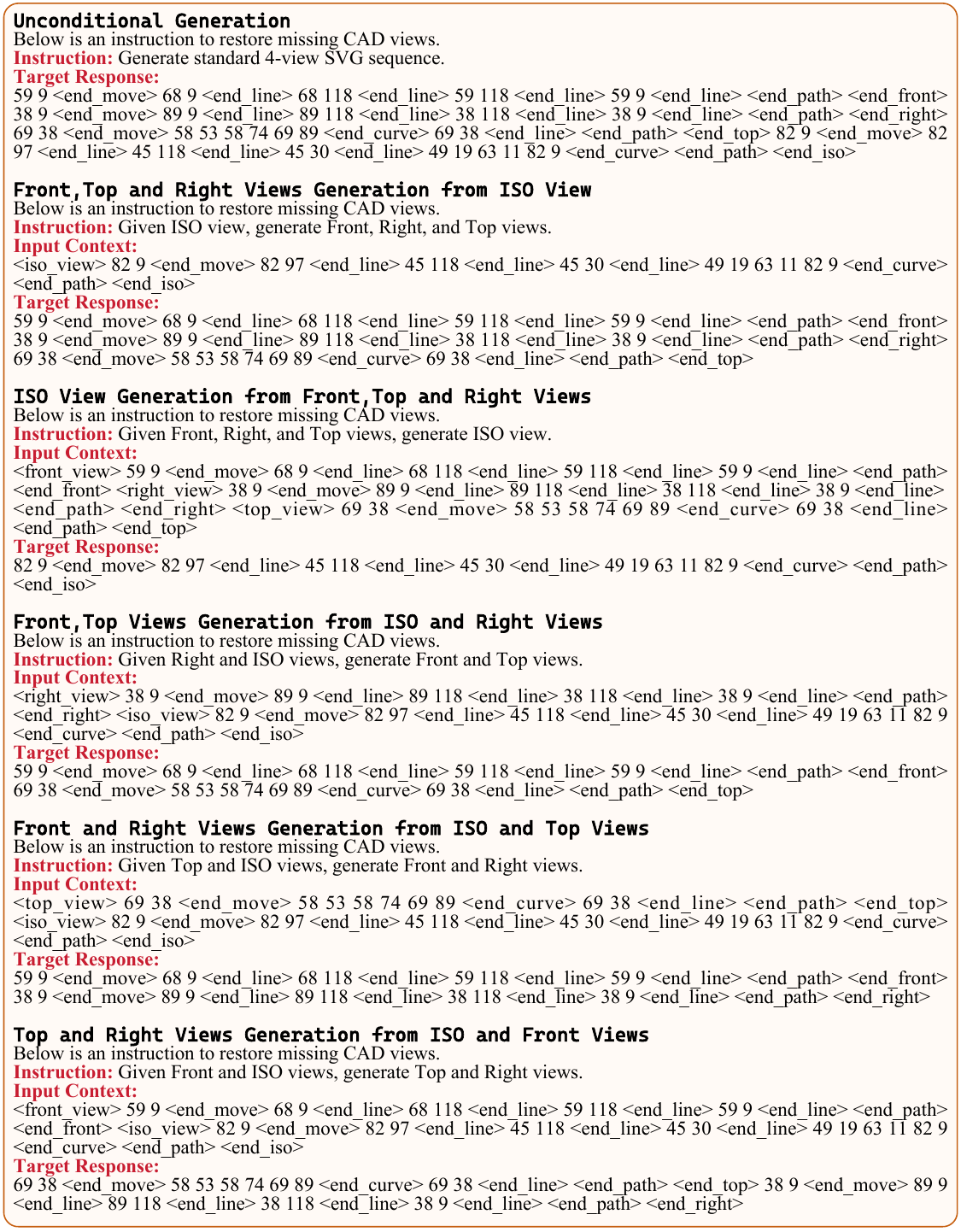}
    \caption{Task-specific instruction templates ($\mathcal{I}$) utilized during the unified multitask training phase.}
    \label{fig:ours_template}
\end{figure}

As introduced in Section \ref{sec:representation}, \textbf{DrawingsDreamer} unifies diverse generation tasks into a single autoregressive framework via task-specific instruction prompts ($\mathcal{I}$). Figure~\ref{fig:ours_template} provides the exact textual instruction templates prepended to the context sequences ($S_{\text{context}}$) for each multitask configuration.

\section{The Impact of Model Capacity}
\label{sec: scale}

To investigate the impact of fundamental Large Language Model capacity on spatial reasoning and multi-view alignment, we extend our framework by scaling \textbf{DrawingsDreamer} up to a 7B parameter architecture (e.g., LLaMA-3-7B). 

\paragraph{Setup.} 
While our primary experiments utilize a 3B parameter model for an optimal balance between performance and computational efficiency, generating highly structured engineering blueprints demands rigorous logical deduction and precise coordinate regression. By scaling up the backbone, we aim to explore whether increased parameter counts can further mitigate the topological hallucinations and dimensional drifts identified in our failure cases (Section~\ref{sec: limit}). The 7B model is fine-tuned using a consistent LoRA configuration ($r=8, \alpha=32$) and identical progressive task-aware curriculum scheduling. Due to memory constraints, the batch size is appropriately adjusted while maintaining the same effective learning rate trajectory.

\paragraph{Quantitative Improvements.} 
Table~\ref{tab:scale_up} summarizes the performance comparison between the 3B and 7B variants. Empirically, scaling up the model yields consistent improvements across all metrics. Most notably, in the Partial View Completion task ($\{v_{\text{iso}}, v_{\text{front}}\} \rightarrow \{v_{\text{top}}, v_{\text{right}}\}$), the 7B model achieves a substantial leap in Parameter Accuracy (from 75.53\% to 82.82\%) and a corresponding drop in Chamfer Distance (from 2.38 to 1.92). This indicates that the larger semantic capacity of the 7B model translates directly into finer coordinate regression and stricter cross-view geometric fidelity.

\begin{table}[htbp]
  \centering
  \caption{Quantitative comparison between 3B and 7B parameter models on Cross-View Translation and Partial View Completion tasks. CD is multiplied by $10^2$. \colorbox{red!20}{Red} and \colorbox{green!20}{green} cells denote the best and second-best results, respectively.}
  \label{tab:scale_up}
  \resizebox{0.9\textwidth}{!}{
  \begin{tabular}{l ccc ccc}
    \toprule
    \multirow{2}{*}{Model} & \multicolumn{3}{c}{$\{v_{\text{iso}}\} \rightarrow \mathcal{V}_{\text{ortho}}$} & \multicolumn{3}{c}{$\{v_{\text{ortho}}\} \rightarrow \mathcal{V}_{\text{iso}}$} \\
    \cmidrule(lr){2-4} \cmidrule(lr){5-7}
    & $Acc_{cmd}\uparrow$ & $Acc_{param}\uparrow$ & CD$\downarrow$ & $Acc_{cmd}\uparrow$ & $Acc_{param}\uparrow$ & CD$\downarrow$ \\
    \midrule
    DrawingsDreamer (3B) & \cellcolor{green!20}89.77\% & \cellcolor{green!20}80.82\% & \cellcolor{green!20}2.41 & \cellcolor{green!20}75.53\% & \cellcolor{green!20}68.36\% & \cellcolor{green!20}2.38 \\
    DrawingsDreamer (7B) & \cellcolor{red!20}91.16\% & \cellcolor{red!20}85.41\% & \cellcolor{red!20}2.04 & \cellcolor{red!20}82.82\% & \cellcolor{red!20}73.51\% & \cellcolor{red!20}1.92  \\
    \bottomrule
  \end{tabular}
  }
\end{table}

\paragraph{Qualitative Observations.} 
Beyond the quantitative metrics, the 7B model demonstrates enhanced robustness against the typical failure modes discussed in Section~\ref{sec: limit}. Specifically, the expanded context-processing capabilities significantly reduce counting errors for repetitive structures (e.g., arrays of circular holes), showcasing stronger local topological memory. Furthermore, the 7B variant exhibits tighter spatial constraints when inferring thin or elongated objects, reducing the coordinate drift that occasionally plagued the 3B model.

\section{Evaluation Metric Details}
\label{sec: eval}
We evaluate \textit{DrawingsDreamer} from two complementary perspectives: symbolic SVG correctness and raster-level visual fidelity. For conditional generation tasks, the model is evaluated only on the views that are required to be generated by the task instruction. For example, in the ISO-to-three-view task, the ISO view is treated as input and excluded from the target set, while the front, right, and top views are evaluated. Conversely, in the three-view-to-ISO task, only the ISO view is evaluated. For tasks conditioned on ISO and one additional orthographic view, the remaining two orthographic views are used as targets.

\paragraph{Symbolic accuracy.}
Each generated SVG sequence and its ground truth sequence are first split into individual views according to the view delimiters \texttt{<end\_front>}, \texttt{<end\_right>}, \texttt{<end\_top>}, and \texttt{<end\_iso>}. Within each target view, we parse the sequence into an ordered list of drawing operations and their numerical parameters. We report two symbolic accuracy metrics. The command accuracy measures whether the generated operation type matches the ground truth at the same sequence position:
\begin{equation}
\mathrm{Acc}_{cmd} = \frac{1}{N}\sum_{i=1}^{N}\mathbbm{1}[c_i = c_i^*],
\end{equation}
where $c_i$ and $c_i^*$ denote the generated and ground-truth operation types, and $N$ is the number of ground-truth operations. The parameter accuracy measures whether numerical parameters are reconstructed within a fixed coordinate tolerance $\tau$:
\begin{equation}
\mathrm{Acc}_{param} =
\mathrm{Acc}_{param} = \frac{1}{M}\sum_i\sum_j \mathbbm{1}\left[|p_{ij}-p_{ij}^*| \leq \tau\right],
\end{equation}
where $p_{ij}$ and $p_{ij}^*$ are generated and ground-truth parameters and $M$ is the number of evaluated ground-truth parameters. Metrics are first averaged over target views for each sample and then averaged across the evaluation set. Missing target views are counted as failed predictions for the corresponding view.

\paragraph{Raster-level visual metrics.}
To evaluate visual fidelity, we render SVG sequences into raster images with a white background at a fixed resolution of $256 \times 256$. All visual metrics are computed per target view and then averaged over the target views of each sample. We report PSNR, SSIM, LPIPS, and Chamfer Distance (CD). PSNR and SSIM measure pixel-level reconstruction quality, while LPIPS measures perceptual similarity using a learned image feature space. All rendered images are normalized to $[0,1]$ before metric computation.

Chamfer Distance is computed on line pixels extracted from the rendered images. We convert each rendered image to grayscale and select foreground pixels with intensity below a fixed threshold. The resulting 2D point sets are normalized by the image resolution. Given the generated point set $P$ and the ground truth point set $Q$, CD is computed as
\begin{equation}
\mathrm{CD}(P,Q) =
\frac{1}{|P|}\sum_{p\in P}\min_{q\in Q}\|p-q\|_2
+
\frac{1}{|Q|}\sum_{q\in Q}\min_{p\in P}\|q-p\|_2 .
\end{equation}
This metric complements raster similarity by directly measuring geometric alignment between predicted and ground-truth line drawings.

\paragraph{Unconditional generation metrics.}
For unconditional generation, there is no one-to-one ground truth target for each generated sample. We therefore evaluate the generated distribution against the test-set distribution using JSD, COV, and MMD. Each four-view SVG is converted into a joint 2D point cloud. Specifically, every view is sampled into a fixed-size point cloud from its line and curve primitives, locally normalized, and placed into a view-specific quadrant. The four view point clouds are then concatenated into a single joint representation.

Let $G$ be the set of generated point clouds and $R$ be the set of reference point clouds. We compute pairwise Chamfer distances between generated and reference samples. Minimum Matching Distance measures the average nearest-neighbor distance from generated samples to the reference set:
\begin{equation}
\mathrm{MMD}(G,R) =
\frac{1}{|G|}\sum_{g\in G}\min_{r\in R}\mathrm{CD}(g,r).
\end{equation}
Coverage measures the fraction of reference samples that are selected as the nearest neighbor of at least one generated sample:
\begin{equation}
\mathrm{COV}(G,R) =
\frac{\left|\left\{\arg\min_{r\in R}\mathrm{CD}(g,r)\;:\;g\in G\right\}\right|}{|R|}.
\end{equation}
MMD reflects sample quality, while COV reflects diversity. We also compute Jensen-Shannon Divergence by voxelizing all points from the generated and reference sets into a fixed 2D occupancy grid and comparing the resulting occupancy distributions:
\begin{equation}
\mathrm{JSD}(P_G,P_R)
= \frac{1}{2}\mathrm{KL}(P_G\|M)
+ \frac{1}{2}\mathrm{KL}(P_R\|M),
\quad
M = \frac{1}{2}(P_G+P_R).
\end{equation}
Lower MMD and JSD indicate better distributional matching, while higher COV indicates better coverage of the test distribution.

\paragraph{Fair comparison with Zero123.}
Zero123 produces raster images rather than SVG command sequences, so symbolic accuracy is not applicable to this baseline. To ensure a fair visual comparison, we evaluate Zero123 using the same raster-level protocol as our SVG outputs. Each predicted target view and its corresponding ground truth view are converted to RGB, composited on a white background when necessary, resized to the same $256 \times 256$ resolution, and evaluated per view. PSNR, SSIM, LPIPS, and CD are then computed with the same definitions and averaged over the same target views. For CD, foreground line pixels are extracted using the same grayscale thresholding rule and the same bidirectional Chamfer formulation. Thus, although Zero123 and DrawingsDreamer use different output representations, all reported visual metrics are computed under an aligned per-view image-space protocol.

\section{Additional Qualitative Results}
\label{sec: add}

\begin{figure}[htbp]
    \centering
    \includegraphics[width=\linewidth]{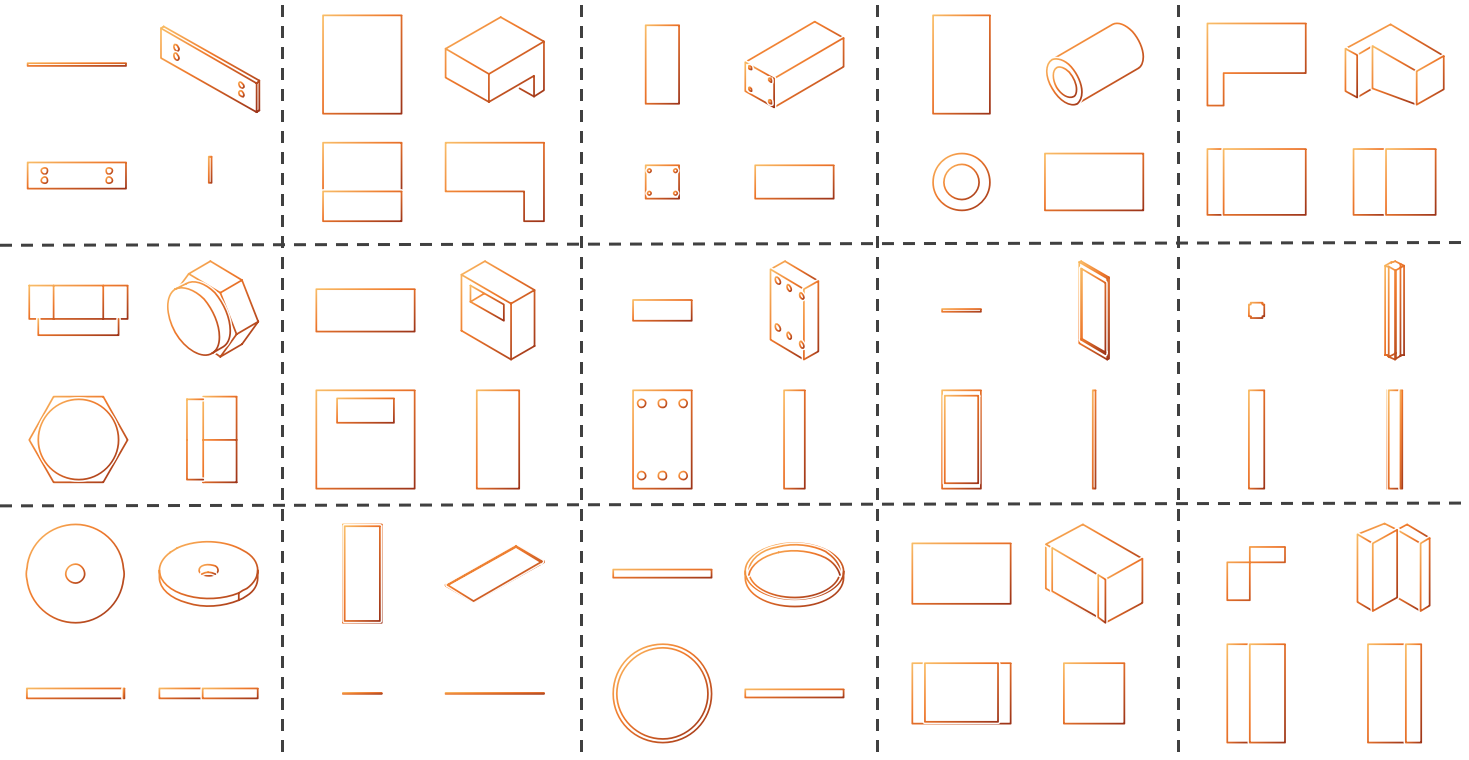}
    \caption{Qualitative results of unconditional multi-view engineering drawing generation.}
    \label{fig:uncond}
\end{figure}

To further demonstrate the generative capacity of \textbf{DrawingsDreamer}, we provide additional qualitative examples of unconditional generation ($\emptyset \rightarrow \mathcal{V}$) in Figure~\ref{fig:uncond}. These results illustrate the model's ability to sample directly from the learned prior distribution, successfully synthesizing diverse, structurally valid, and strictly aligned multi-view engineering blueprints without any contextual conditioning.

\section{Limitations and Future Work}
\label{sec: limit}

\begin{figure}[htbp]
    \centering
    \includegraphics[width=\linewidth]{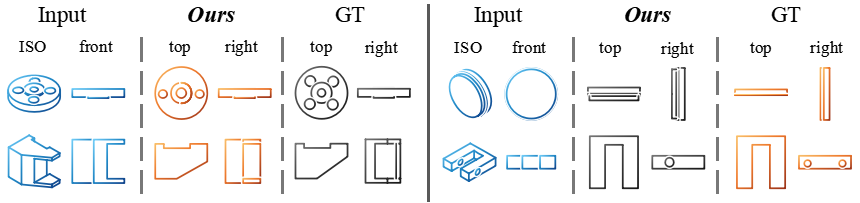}
    \caption{Failure cases  during Partial View Completion.}
    \label{fig:fail_top_right}
\end{figure}

\begin{figure}[htbp]
    \centering
    \includegraphics[width=\linewidth]{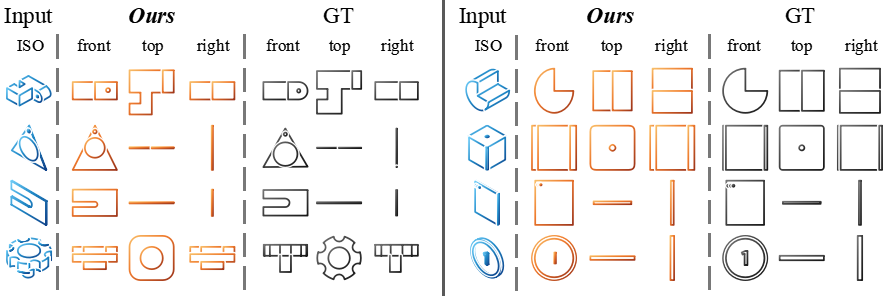}
    \caption{Failure case  during Isometric to Orthographic translation.}
    \label{fig:fail_rest}
\end{figure}

While \textbf{DrawingsDreamer} demonstrates strong capabilities in generating multi-view engineering drawings, we observe certain limitations, primarily manifesting in two distinct failure modes as illustrated in Figures~\ref{fig:fail_top_right}, \ref{fig:fail_rest}.

\paragraph{Shape and Topological Reasoning Errors.} 
The first category of failure involves structural hallucinations, most commonly taking the form of incorrect inference regarding discrete geometric features. For instance, the model may occasionally generate an incorrect number of repeating elements, such as missing a circular hole or hallucinating an extra symmetric slot. We attribute this to the known limitations of purely autoregressive models when handling highly repetitive structural tokens over extended context windows. Without explicit counting mechanisms, the attention mechanism can occasionally lose track of repeating sub-sequences.

\paragraph{Geometric and Dimensional Errors.} 
The second category pertains to precise spatial reasoning, which is particularly pronounced when the model attempts to generate flat, thin, or highly elongated objects. In these scenarios, the model can struggle to maintain rigorous dimensional consistency across multiple views, resulting in inaccurate lengths, irregular thicknesses, distorted hole radii, or misaligned spatial coordinates. This issue stems from the inherent challenges of mapping continuous, highly skewed geometric spaces into discrete token vocabularies. For extremely thin features, a minor coordinate drift or quantization artifact in one projection can be amplified during cross-view translation, compromising the strict spatial alignment required for professional CAD blueprints.

\paragraph{Future Work.} 
Addressing these limitations presents straightforward avenues for future research. While introducing explicit cross-view alignment rules could provide immediate structural constraints, a more scalable approach involves Reinforcement Learning (RL). By designing reward models that explicitly penalize cross-view geometric discrepancies, the framework can be optimized to internalize strict spatial alignment, preserving a purely data-driven sequence modeling paradigm without relying on hand-crafted heuristics.

\begin{figure}[htbp]
    \centering
    \includegraphics[width=0.8\linewidth]{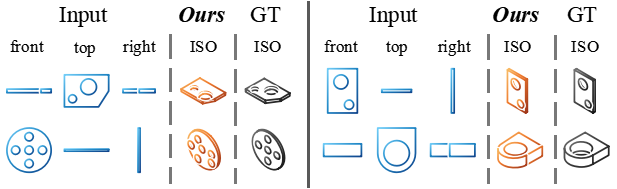}
    \caption{Failure case during Orthographic to Isometric reconstruction.}
    \label{fig:iso}
\end{figure}

\newpage
\section*{NeurIPS Paper Checklist}

\begin{enumerate}

\item {\bf Claims}
    \item[] Question: Do the main claims made in the abstract and introduction accurately reflect the paper's contributions and scope?
    \item[] Answer: \answerYes{} % Replace by \answerYes{}, \answerNo{}, or \answerNA{}.
    \item[] Justification: The main claims made in the abstract and introduction accurately reflect the paper’s contributions and scope.

\item {\bf Limitations}
    \item[] Question: Does the paper discuss the limitations of the work performed by the authors?
    \item[] Answer: \answerYes{} % Replace by \answerYes{}, \answerNo{}, or \answerNA{}.
    \item[] Justification: We mention our limitations in Appendix \ref{sec: limit}.
    \item[] Guidelines:
    \begin{itemize}
        \item The answer NA means that the paper has no limitation while the answer No means that the paper has limitations, but those are not discussed in the paper. 
        \item The authors are encouraged to create a separate ``Limitations'' section in their paper.
        \item The paper should point out any strong assumptions and how robust the results are to violations of these assumptions (e.g., independence assumptions, noiseless settings, model well-specification, asymptotic approximations only holding locally). The authors should reflect on how these assumptions might be violated in practice and what the implications would be.
        \item The authors should reflect on the scope of the claims made, e.g., if the approach was only tested on a few datasets or with a few runs. In general, empirical results often depend on implicit assumptions, which should be articulated.
        \item The authors should reflect on the factors that influence the performance of the approach. For example, a facial recognition algorithm may perform poorly when image resolution is low or images are taken in low lighting. Or a speech-to-text system might not be used reliably to provide closed captions for online lectures because it fails to handle technical jargon.
        \item The authors should discuss the computational efficiency of the proposed algorithms and how they scale with dataset size.
        \item If applicable, the authors should discuss possible limitations of their approach to address problems of privacy and fairness.
        \item While the authors might fear that complete honesty about limitations might be used by reviewers as grounds for rejection, a worse outcome might be that reviewers discover limitations that aren't acknowledged in the paper. The authors should use their best judgment and recognize that individual actions in favor of transparency play an important role in developing norms that preserve the integrity of the community. Reviewers will be specifically instructed to not penalize honesty concerning limitations.
    \end{itemize}

\item {\bf Theory assumptions and proofs}
    \item[] Question: For each theoretical result, does the paper provide the full set of assumptions and a complete (and correct) proof?
    \item[] Answer: \answerNA{} % Replace by \answerYes{}, \answerNo{}, or \answerNA{}.
    \item[] Justification: The paper does not include theoretical results.

    \item {\bf Experimental result reproducibility}
    \item[] Question: Does the paper fully disclose all the information needed to reproduce the main experimental results of the paper to the extent that it affects the main claims and/or conclusions of the paper (regardless of whether the code and data are provided or not)?
    \item[] Answer: \answerYes{} % Replace by \answerYes{}, \answerNo{}, or \answerNA{}.
    \item[] Justification: We disclose all information needed for reproduction in Section \ref{sec: experiments}.
    \item[] Guidelines:
    \begin{itemize}
        \item The answer NA means that the paper does not include experiments.
        \item If the paper includes experiments, a No answer to this question will not be perceived well by the reviewers: Making the paper reproducible is important, regardless of whether the code and data are provided or not.
        \item If the contribution is a dataset and\slash or model, the authors should describe the steps taken to make their results reproducible or verifiable. 
        \item Depending on the contribution, reproducibility can be accomplished in various ways. For example, if the contribution is a novel architecture, describing the architecture fully might suffice, or if the contribution is a specific model and empirical evaluation, it may be necessary to either make it possible for others to replicate the model with the same dataset, or provide access to the model. In general. releasing code and data is often one good way to accomplish this, but reproducibility can also be provided via detailed instructions for how to replicate the results, access to a hosted model (e.g., in the case of a large language model), releasing of a model checkpoint, or other means that are appropriate to the research performed.
        \item While NeurIPS does not require releasing code, the conference does require all submissions to provide some reasonable avenue for reproducibility, which may depend on the nature of the contribution. For example
        \begin{enumerate}
            \item If the contribution is primarily a new algorithm, the paper should make it clear how to reproduce that algorithm.
            \item If the contribution is primarily a new model architecture, the paper should describe the architecture clearly and fully.
            \item If the contribution is a new model (e.g., a large language model), then there should either be a way to access this model for reproducing the results or a way to reproduce the model (e.g., with an open-source dataset or instructions for how to construct the dataset).
            \item We recognize that reproducibility may be tricky in some cases, in which case authors are welcome to describe the particular way they provide for reproducibility. In the case of closed-source models, it may be that access to the model is limited in some way (e.g., to registered users), but it should be possible for other researchers to have some path to reproducing or verifying the results.
        \end{enumerate}
    \end{itemize}

\item {\bf Open access to data and code}
    \item[] Question: Does the paper provide open access to the data and code, with sufficient instructions to faithfully reproduce the main experimental results, as described in supplemental material?
    \item[] Answer: \answerYes{} % Replace by \answerYes{}, \answerNo{}, or \answerNA{}.
    \item[] Justification: We will provide data, code, and instructions for the final paper.
    \item[] Guidelines:
    \begin{itemize}
        \item The answer NA means that paper does not include experiments requiring code.
        \item Please see the NeurIPS code and data submission guidelines (\url{https://neurips.cc/public/guides/CodeSubmissionPolicy}) for more details.
        \item While we encourage the release of code and data, we understand that this might not be possible, so "No" is an acceptable answer. Papers cannot be rejected simply for not including code, unless this is central to the contribution (e.g., for a new open-source benchmark).
        \item The instructions should contain the exact command and environment needed to run to reproduce the results. See the NeurIPS code and data submission guidelines (\url{https://neurips.cc/public/guides/CodeSubmissionPolicy}) for more details.
        \item The authors should provide instructions on data access and preparation, including how to access the raw data, preprocessed data, intermediate data, and generated data, etc.
        \item The authors should provide scripts to reproduce all experimental results for the new proposed method and baselines. If only a subset of experiments are reproducible, they should state which ones are omitted from the script and why.
        \item At submission time, to preserve anonymity, the authors should release anonymized versions (if applicable).
        \item Providing as much information as possible in supplemental material (appended to the paper) is recommended, but including URLs to data and code is permitted.
    \end{itemize}

\item {\bf Experimental setting/details}
    \item[] Question: Does the paper specify all the training and test details (e.g., data splits, hyperparameters, how they were chosen, type of optimizer) necessary to understand the results?
    \item[] Answer: \answerYes{} % Replace by \answerYes{}, \answerNo{}, or \answerNA{}.
    \item[] Justification: We specifies all the training and test details in Section \ref{sec: experiments}.

\item {\bf Experiment statistical significance}
    \item[] Question: Does the paper report error bars suitably and correctly defined or other appropriate information about the statistical significance of the experiments?
    \item[] Answer: \answerYes{} % Replace by \answerYes{}, \answerNo{}, or \answerNA{}.
    \item[] Justification: We report consistent performance across multiple runs and use fixed random seed settings to support the statistical significance of our results.
    \item[] Guidelines:
    \begin{itemize}
        \item The answer NA means that the paper does not include experiments.
        \item The authors should answer "Yes" if the results are accompanied by error bars, confidence intervals, or statistical significance tests, at least for the experiments that support the main claims of the paper.
        \item The factors of variability that the error bars are capturing should be clearly stated (for example, train/test split, initialization, random drawing of some parameter, or overall run with given experimental conditions).
        \item The method for calculating the error bars should be explained (closed form formula, call to a library function, bootstrap, etc.)
        \item The assumptions made should be given (e.g., Normally distributed errors).
        \item It should be clear whether the error bar is the standard deviation or the standard error of the mean.
        \item It is OK to report 1-sigma error bars, but one should state it. The authors should preferably report a 2-sigma error bar than state that they have a 96\% CI, if the hypothesis of Normality of errors is not verified.
        \item For asymmetric distributions, the authors should be careful not to show in tables or figures symmetric error bars that would yield results that are out of range (e.g., negative error rates).
        \item If error bars are reported in tables or plots, the authors should explain in the text how they were calculated and reference the corresponding figures or tables in the text.
    \end{itemize}

\item {\bf Experiments compute resources}
    \item[] Question: For each experiment, does the paper provide sufficient information on the computer resources (type of compute workers, memory, time of execution) needed to reproduce the experiments?
    \item[] Answer: \answerYes{} % Replace by \answerYes{}, \answerNo{}, or \answerNA{}.
    \item[] Justification: We provide information about the computer resources used in Section \ref{sec: setup}.
    
\item {\bf Code of ethics}
    \item[] Question: Does the research conducted in the paper conform, in every respect, with the NeurIPS Code of Ethics \url{https://neurips.cc/public/EthicsGuidelines}?
    \item[] Answer: \answerYes{} % Replace by \answerYes{}, \answerNo{}, or \answerNA{}.
    \item[] Justification: We have read and understood the code of ethics and have made every effort to adhere to it.

\item {\bf Broader impacts}
    \item[] Question: Does the paper discuss both potential positive societal impacts and negative societal impacts of the work performed?
    \item[] Answer: \answerNA{} % Replace by \answerYes{}, \answerNo{}, or \answerNA{}.
    \item[] Justification: Our work contributes to Engineering Drawings Generation. It does not impact society at large.
    
\item {\bf Safeguards}
    \item[] Question: Does the paper describe safeguards that have been put in place for responsible release of data or models that have a high risk for misuse (e.g., pre-trained language models, image generators, or scraped datasets)?
    \item[] Answer: \answerNA{} % Replace by \answerYes{}, \answerNo{}, or \answerNA{}.
    \item[] Justification: The paper poses no such risks.

\item {\bf Licenses for existing assets}
    \item[] Question: Are the creators or original owners of assets (e.g., code, data, models), used in the paper, properly credited and are the license and terms of use explicitly mentioned and properly respected?
    \item[] Answer: \answerYes{} % Replace by \answerYes{}, \answerNo{}, or \answerNA{}.
    \item[] Justification: All existing assets, including code, data, and models, are properly cited in the paper, and their licenses and usage terms are respected in accordance with the original sources.

\item {\bf New assets}
    \item[] Question: Are new assets introduced in the paper well documented and is the documentation provided alongside the assets?
    \item[] Answer: \answerYes{} % Replace by \answerYes{}, \answerNo{}, or \answerNA{}.
    \item[] Justification: We will release our codebase along with detailed README files to ensure usability and reproducibility in an anonymous manner.
    \item[] Guidelines:
    \begin{itemize}
        \item The answer NA means that the paper does not release new assets.
        \item Researchers should communicate the details of the dataset\slash code\slash model as part of their submissions via structured templates. This includes details about training, license, limitations, etc. 
        \item The paper should discuss whether and how consent was obtained from people whose asset is used.
        \item At submission time, remember to anonymize your assets (if applicable). You can either create an anonymized URL or include an anonymized zip file.
    \end{itemize}

\item {\bf Crowdsourcing and research with human subjects}
    \item[] Question: For crowdsourcing experiments and research with human subjects, does the paper include the full text of instructions given to participants and screenshots, if applicable, as well as details about compensation (if any)? 
    \item[] Answer: \answerNA{}% Replace by \answerYes{}, \answerNo{}, or \answerNA{}.
    \item[] Justification: This work does not use crowdsourcing or human subjects.

\item {\bf Institutional review board (IRB) approvals or equivalent for research with human subjects}
    \item[] Question: Does the paper describe potential risks incurred by study participants, whether such risks were disclosed to the subjects, and whether Institutional Review Board (IRB) approvals (or an equivalent approval/review based on the requirements of your country or institution) were obtained?
    \item[] Answer: \answerNA{} % Replace by \answerYes{}, \answerNo{}, or \answerNA{}.
    \item[] Justification: This work does not use crowdsourcing or human subjects.

\item {\bf Declaration of LLM usage}
    \item[] Question: Does the paper describe the usage of LLMs if it is an important, original, or non-standard component of the core methods in this research? Note that if the LLM is used only for writing, editing, or formatting purposes and does \emph{not} impact the core methodology, scientific rigor, or originality of the research, declaration is not required.
    %this research? 
    \item[] Answer: \answerYes{} % Replace by \answerYes{}, \answerNo{}, or \answerNA{}.
    \item[] Justification: We describe the usage of LLMs in Section \ref{sec: method} and details in Section \ref{sec: setup}.
    \item[] Guidelines:
    \begin{itemize}
        \item The answer NA means that the core method development in this research does not involve LLMs as any important, original, or non-standard components.
        \item Please refer to our LLM policy in the NeurIPS handbook for what should or should not be described.
    \end{itemize}

\end{enumerate}

\end{document}